\documentclass{article} % For LaTeX2e

\newif\ifneurips
\neuripsfalse

\newif\ificlrpreprint
\iclrpreprinttrue
\ifneurips
  \PassOptionsToPackage{numbers, sort&compress}{natbib}
  \usepackage[dblblindworkshop, final]{neurips_2026}   % or sglblindworkshop
\workshoptitle{Interpretability as a Science}

\else
  \usepackage{iclr2027_conference,times}
\fi

\usepackage{amsmath,amsfonts,bm}

\def\eqref#1{equation~\ref{#1}}
\def\1{\bm{1}}

\DeclareMathAlphabet{\mathsfit}{\encodingdefault}{\sfdefault}{m}{sl}
\SetMathAlphabet{\mathsfit}{bold}{\encodingdefault}{\sfdefault}{bx}{n}

\usepackage{hyperref}
\usepackage{url}
\usepackage{cleveref}
\usepackage{enumitem}
\usepackage{fontawesome5}
\AddToHook{cmd/appendix/before}{%
  \crefalias{section}{appendix}%
  \crefalias{subsection}{appendix}%
}
\usepackage{appendix}

\usepackage[T1]{fontenc}
\usepackage[utf8]{inputenc}
\usepackage{microtype}
\usepackage{graphicx}

\usepackage{amssymb}
\usepackage{mathtools}

\usepackage{booktabs}
\usepackage{multirow}
\usepackage[table]{xcolor}
\usepackage{tabularx}
\usepackage{siunitx}
\usepackage{wrapfig}

\newcommand{\myparagraph}[1]{\textbf{#1}\hspace{0.1em}}

\newif\ifshowrebuttal
\showrebuttalfalse
\definecolor{rebuttalcolor}{RGB}{112,48,160}
\newcommand{\rebuttal}[1]{\ifshowrebuttal{\color{rebuttalcolor}#1}\else#1\fi}
\newenvironment{rebuttalblock}{\ifshowrebuttal\color{rebuttalcolor}\fi}{}

\title{Persistent Sparse Autoencoders: \\Learning Feature-Specific Timescales in Language Model Representations}
\author{%
  Haoyan Luo \\
  University of Cambridge \\
  \And
  Mateo Espinosa Zarlenga \\
  University of Oxford \\
  \And
  Mateja Jamnik \\
  University of Cambridge \\
}

\ifneurips\else\ificlrpreprint\iclrfinalcopy\fi\fi
\begin{document}

\maketitle
\ifneurips\else\ificlrpreprint\lhead{Preprint}\fi\fi

\ifneurips\else\ificlrpreprint
  \vspace{-1.5em}
  \begin{center}
    {\hypersetup{hidelinks}%
    \faGithub\ \href{https://github.com/clarenceluo78/persistent-sparse-autoencoders}{\textcolor[RGB]{120,60,104}{\texttt{\textbf{Code}}}}%
    \hspace{2.5em}%
    \raisebox{-0.15em}{\includegraphics[height=1em]{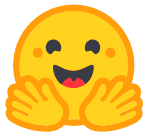}}\ \href{https://huggingface.co/haoyanl78/persistent-sparse-autoencoders}{\textcolor[RGB]{120,60,104}{\texttt{\textbf{HuggingFace}}}}}
  \end{center}
  \vspace{0.8em}
\fi\fi

\begin{abstract}
Sparse autoencoders (SAEs) decompose language model activations into sparse features, yet these models traditionally encode each token independently, failing to expose information that persists across a sequence. We first show that temporal persistence can naturally emerge in standard SAE features: after a feature activates, the hidden state remains aligned with its direction, and past activations reduce the reconstruction error of later hidden states. How long this lasts varies widely across features.
We therefore introduce \emph{Persistent Sparse Autoencoders} (Persistent SAEs), an extension of standard SAEs that learns a persistence coefficient for each feature, allowing feature-specific timescales to emerge from reconstruction.
% without deciding in advance which features should carry information forward.
Our experiments show that Persistent SAEs retain competitive reconstruction quality while learning a spectrum of timescales: short-timescale (fast) features remain locally interpretable, whereas long-timescale (slow) features accumulate higher-level contextual information. 
Moreover, we show in a prompt-injection monitoring case study that slow features retain injection signals over long contexts and causally affect monitor judgements. 
% These results show that feature timescale provides an unsupervised axis for organising sparse representations and exposes persistent state useful for interpreting and monitoring language models.\looseness-1
These results suggest that Persistent SAEs offer new opportunities for interpreting and monitoring language models via persistent features.\looseness-1
\end{abstract}

\section{Introduction}

% \begin{figure}[t]
%     \centering
%     \includegraphics[width=\linewidth]{figures/teaser.pdf}
%     \caption{
%     \textbf{From token-wise activations to a persistent state.} Persistent SAEs learn a timescale for each feature: fast features remain token-specific and locally interpretable, while slow features integrate sparse activations over tokens to track the active context.
%     }
%     \label{fig:teaser}
% \end{figure}

Understanding the output of a large language model (LLM) requires identifying what information is represented in its hidden states and how that information evolves throughout the sequence \citep{from_understanding_to_utilization_survey, sharkey2025open}.
This is challenging because hidden states are high-dimensional representations in which the encoded information is not directly interpretable.
Sparse autoencoders (SAEs) \citep{huben2024sparse, lan2024sparse} are methods that decompose an LLM's hidden states into a set of sparse \emph{features}, and have become a popular tool for approaching this problem. Specifically, when features are aligned with human-interpretable concepts, an SAE's decomposition enables easy inspection of concepts learnt by an LLM and provides a mechanism for model control through the inspection of and intervention on these features \citep{templeton2024scaling,marks2025sparse,sae_right_features_steering}.\looseness=-1

Although causal transformer hidden states integrate the preceding context, standard SAEs encode each token independently \citep{huben2024sparse,batchtopk_sae,matryoshka_sae}. This atemporal formulation favours token-specific features but obscures the contextual information a model carries across a sequence, such as the current topic, style or task state \citep{temporal_sparse_autoencoders}.
To alleviate this constraint, recent methods have incorporated temporal structure into SAEs through predefined partitions and decompositions of temporal roles \citep{temporal_sparse_autoencoders,priors_in_time}.
However, these approaches require specifying, a priori, which features should carry temporal information, leading to experimenter bias in the features the SAE learns.
% It therefore remains unclear whether the temporality of features can be learnt directly from the data, without the need to hardcode a feature's when building the model.
It therefore remains unclear whether each feature's timescale can be learned directly from the data rather than being predefined by the model architecture.\looseness=-1

This paper explores this question starting from the observation that temporal persistence can naturally emerge in standard SAE features. As shown in \Cref{fig:main-observation}, we observe that, after a feature activates, later hidden states remain more strongly aligned with its decoder direction, and past activations contain information that helps reconstruct subsequent hidden states. 
The persistence decays at substantially different rates across features, suggesting that different features operate at different timescales.
Yet, standard SAEs encode each position independently and therefore do not explicitly represent this persistence.
To fill this gap, we introduce \emph{Persistent Sparse Autoencoders}, which learn a persistence coefficient $\lambda$ for every feature. Each feature keeps a persistent \emph{state} that decays according to its persistence and is updated by a token-wise sparse activation.
If carrying a feature's state forward helps reconstruct later hidden states, there is an incentive to learn a higher persistence during optimisation; if it does not, the objective favours a smaller persistence and faster decay.
% We do not assign roles to timescales in advance: reconstruction sets a timescale for every feature, and we then ask what is represented at each.

%First figure here?

Our experiments show that Persistent SAEs retain competitive reconstruction quality while learning a broad range of feature timescales.
Most features are \emph{fast features} that behave as locally interpretable detectors, as in a standard SAE. \emph{Slow features}, in contrast, accumulate the higher-level contextual information: selecting features by learned persistence requires no labels, yet a linear probe using the slowest features predicts the current subject nearly as well as one using features selected with subject labels.
We show that, in a prompt-injection monitoring case study \citep{agentdojo}, slow features keep an injection detectable, and are causally effective on a monitor's judgement over long contexts.
% In a prompt-injection monitoring case study \citep{agentdojo}, these persistent states keep an injection detectable, and keep the corresponding direction causally effective on a monitor's judgement, hundreds of tokens after the injected text has passed.
% These results show that Persistent SAEs can organise sparse features by timescale, separating token-wise activations from a persistent state.
Our main contributions can be summarised as follows:
\begin{itemize}[leftmargin=9pt,topsep=0pt,itemsep=-1pt]
    \item We show that SAE features exhibit temporal persistence: after activation, the signal along a feature's decoder direction remains elevated, and past activations help reconstruct later hidden states.

    \item We introduce \emph{Persistent Sparse Autoencoders}, an SAE variant that models temporal persistence with learnable feature-wise persistence coefficients. Persistent SAEs retain competitive reconstruction quality while discovering a broad spectrum of feature timescales without predefined temporal roles.

    \item We show that short-timescale (fast) features act as locally interpretable detectors, while long-timescale (slow) features retain contextual information across a sequence. A prompt-injection monitoring case study demonstrates the potential of slow features for long-context monitoring and intervention, opening new opportunities to interpret and control LLMs via explicitly persistent features.\looseness=-1
\end{itemize}
\section{Background and Related Work}

\myparagraph{Sparse autoencoders.}
Given an LLM's hidden state $\mathbf{x}^{(t)} \in \mathbb{R}^d$ at token $t$, a standard SAE \citep{huben2024sparse} encodes $\mathbf{x}^{(t)}$ as non-negative sparse activations $\mathbf{f}^{(t)} \in \mathbb{R}_{\geq 0}^m$ from which it can be reconstructed as
\begin{equation}
    \hat{\mathbf{x}}^{(t)}_{\mathrm{base}}
    = \mathbf{W}_{\mathrm{dec}} \mathbf{f}^{(t)} + \mathbf{b}_{\mathrm{dec}}.
    \label{eq:standard-reconstruction}
\end{equation}
Under this representation, the $j$-th \emph{feature} is represented by the activation $\mathbf{f}^{(t)}_j$ and its corresponding normalised decoder direction $\mathbf{d}_j=\mathbf{W}_{\mathrm{dec}}[:,j]/\lVert \mathbf{W}_{\mathrm{dec}}[:,j]\rVert_2$ (see \Cref{app:notation} for a notation summary). Intuitively, we think of feature $j$ as \emph{active} when $\mathbf{f}^{(t)}_j>0$.
Because $\mathbf{f}^{(t)}$ is sparse by construction, each reconstruction depends on only a small set of features, making the contribution of individual features easier to inspect \citep{huben2024sparse,bricken2023monosemanticity}.
% Subsequent SAEs, such as BatchTopK \citep{batchtopk_sae} and hierarchical Matryoshka SAEs \citep{matryoshka_sae}, improve how a representation's capacity is allocated and organised.
Subsequent SAEs improve how representational capacity is allocated and organised: BatchTopK SAEs \citep{batchtopk_sae} replace the fixed per-token TopK constraint with a batch-level sparsity budget, allowing the number of active features to adapt across tokens, while Matryoshka SAEs \citep{matryoshka_sae} jointly train nested dictionaries of increasing size, encouraging smaller dictionaries to learn general features while additional capacity in larger dictionaries captures more specific ones.

\myparagraph{Temporal structure in LLM representations.}
Although autoregressive LLMs build $\mathbf{x}^{(t)}$ from the preceding context, standard SAEs and most SAE variants compute features $\mathbf{f}^{(t)}$ independently of the features observed in preceding tokens. SAE features, therefore, do not explicitly expose information carried across tokens. Yet, traditional slow feature analysis shows that slowly varying latent factors reveal higher-level features in neural networks \citep{foldiak1991learning,wiskott2002slow}.
% and cortical recordings show a hierarchy of intrinsic timescales, not a single shared rate \citep{hasson2008hierarchy,murray2014hierarchy}. 
% Sequence models show the same heterogeneity: a unit's timescale in an LSTM language model is set by its forget-gate bias, and trained models learn a heavy-tailed distribution of these timescales \citep{tallec2018warp,mahto2021multi}, mirroring the power-law decay of dependencies in text \citep{lin2017critical}. 
Prior work likewise shows that language models track evolving entity states \citep{kim2023entity}, while their internal representations encode game-board states \citep{li2023emergent}, spatial and temporal coordinates \citep{gurnee2024space}, and propositional world states \citep{feng2025monitoring}. This motivates asking whether such state information persists across tokens and can be explicitly captured by SAE features.\looseness=-1
% Prior studies likewise show that language models track entities and that their hidden states encode game-board, spatial, temporal, and propositional states \citep{kim2023entity,li2023emergent,gurnee2024space,feng2025monitoring}. These findings motivate mechanisms that explicitly encode temporal persistence in SAE features.

\myparagraph{Temporal extensions of SAEs.}
Recent temporal extensions of SAEs \citep{temporal_sparse_autoencoders, priors_in_time} have made progress toward capturing feature persistence. Temporal SAEs \citep{temporal_sparse_autoencoders} pre-assign a high-level feature subset and use a contrastive objective to align their activations across nearby tokens. \citet{priors_in_time} instead use an attention-based predictor to separate context-predictable information from a sparse novel residual representation. 
However, both approaches impose temporal structure through a predefined partition of the feature space. In contrast, we ask whether each feature's temporal persistence can be learned directly from the data.
If SAE features differ in temporal persistence, we could model this variation during training, giving each feature an explicit timescale for inspection and analysis.%\looseness=-1
% However, both methods decide before training which features carry temporal information, by fixing a partition. 
% Both methods prescribe temporal structure via a fixed partition. However, neither allows each sparse feature to learn its own temporal persistence, complicating post-hoc analysis of the expected timescale of different features. 
% Persistent SAEs fill this gap by assigning each feature a persistent state and a learned persistence coefficient that can be easily analysed.

\section{Does Temporal Structure Exist at the Feature Level?}
\label{sec:temporal-persistence}

We first investigate whether SAEs inherit temporal structure from the hidden states they reconstruct via two experiments: (1)~whether the signal along $\mathbf{d}_j$ persists after feature $j$ activates, and (2)~whether past activations improve the reconstruction of later hidden states.

\myparagraph{Setup.}
We analyse residual-stream hidden states $\{\mathbf{x}^{(t)}\}_{t=1}^T$ from Pythia-160M-deduped (layer~8) \citep{biderman2023pythia} and Gemma-2-2B (layer~12) \citep{gemma2} on Pile documents \citep{gao2020pile}. We train BatchTopK \citep{batchtopk_sae} and Matryoshka \citep{matryoshka_sae} SAEs, neither of which has a temporal prior. We describe training configurations in \Cref{app:training-details}.

\begin{figure*}[!t]
    \centering
    \includegraphics[width=\linewidth]{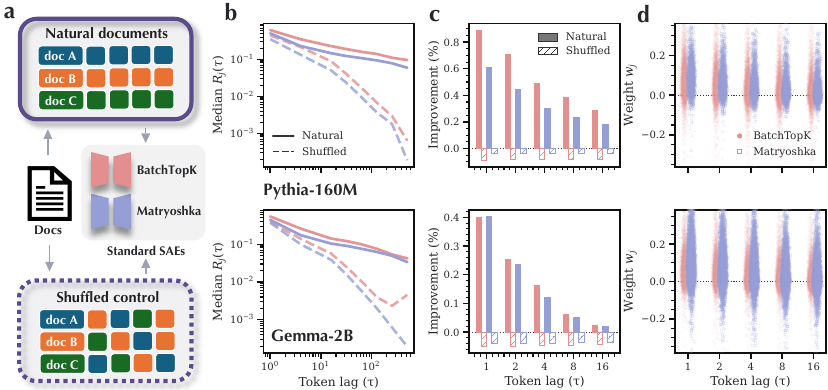}
    \caption{
    \textbf{Standard SAE features exhibit temporal structure that is not explicitly modelled.}
    \textbf{(a)}~Natural Pile sequences preserve local and global continuity, whereas shuffled control sequences preserve local continuity while breaking global continuity.
    \textbf{(b)}~Empirical feature persistence $R_j(\tau)$ decays more slowly in natural sequences than in shuffled sequences.
    \textbf{(c)}~Past feature activations improve held-out reconstruction of later hidden states only in natural sequences.
    \textbf{(d)}~Fitted context-correction weights \(w_j\) are predominantly positive across token lags, indicating that past feature activations continue to contribute along the same feature directions.
    % At lag $\tau=1$, fitted context-correction weights $w_j$ are predominantly positive for natural sequences but centred near zero for shuffled ones.
    }
    \label{fig:main-observation}
\end{figure*}

Natural Pile documents are continuous at two scales: tokens form coherent sentences (local continuity), and the sentences collectively develop a coherent document-level context (global continuity). Our \emph{shuffled control} isolates the contribution of global continuity by concatenating sentences drawn from different Pile documents (\Cref{fig:main-observation}a). 
% Natural documents and the shuffled control therefore both preserve local continuity, but only natural documents preserve global continuity. 
Comparing the two lets us test whether feature persistence depends on document-level continuity, beyond the local continuity preserved within individual sentences.
% Comparing them lets us test whether feature persistence reflects connections across the document beyond the local coherence within individual sentences. 
\Cref{app:dataset-details} gives the construction details of the dataset.\looseness=-1

\myparagraph{Sparse features temporally persist in the hidden states.}
We first test whether the signal aligned with feature $j$ persists after it activates. At each token position $t$, we define $p^{(t)}_j = \mathbf{d}_j^\top \mathbf{x}^{(t)}$ as the projection of the hidden state $\mathbf{x}^{(t)}$ for the $t$-th token onto feature $j$'s decoder  direction.
Intuitively, a larger $p^{(t)}_j$ indicates a stronger alignment between the hidden state and feature direction $\mathbf{d}_j$. 
% (and, therefore, a higher likelihood that feature $j$ is active).
Conditioning on feature $j$ being active at token $t$, we then define its \emph{empirical persistence} at lag $\tau$, where $\tau$ is the number of subsequent tokens, as
\begin{equation}
    R_j(\tau) :=
    \big({
    \mathbb{E}\!\left[p^{(t+\tau)}_j \mid \mathbf{f}^{(t)}_j > 0\right]
    - \mathbb{E}\!\left[p^{(t)}_j\right]
    }\Big) \Big/ \Big({
    \mathbb{E}\!\left[p^{(t)}_j \mid \mathbf{f}^{(t)}_j > 0\right]
    - \mathbb{E}\!\left[p^{(t)}_j\right]
    }\Big),
    \label{eq:persistence}
\end{equation}
% \begin{equation}
%     R_j(\tau) :=
%     \frac{
%     \mathbb{E}\!\left[p^{(t+\tau)}_j \mid \mathbf{f}^{(t)}_j > 0\right]
%     - \mathbb{E}\!\left[p^{(t)}_j\right]
%     }{
%     \mathbb{E}\!\left[p^{(t)}_j \mid \mathbf{f}^{(t)}_j > 0\right]
%     - \mathbb{E}\!\left[p^{(t)}_j\right]
%     },
%     \label{eq:persistence}
% \end{equation}
where $\mathbb{E}[p^{(t)}_j]$ is the feature's average projection across all evaluated tokens. 
A larger $R_j(\tau)$ indicates greater persistence: $R_j(\tau)=1$ means that the increase in feature-aligned projection at activation is fully retained after $\tau$ tokens, while $R_j(\tau)=0$ means that the projection has returned to baseline. \Cref{app:persistence-derivation} provides the evaluation procedure and feature-selection criteria.

In \Cref{fig:main-observation}b, we observe that across both models and SAE families, empirical feature persistence decays more slowly in natural documents than in the shuffled control. This suggests that standard SAE features inherit the persistence associated with global continuity, even when the SAE does not explicitly encode it.
These results suggest that feature-aligned signals persist beyond the tokens that activate the feature. Therefore, it may be possible to model this temporal structure explicitly and test whether past activations help reconstruct the hidden states of subsequent tokens.

\myparagraph{Past feature activations can help reconstruction.}
% Persistence alone does not show that past activations contain information useful for reconstruction. 
We further test whether $\mathbf{f}^{(t-\tau)}$ can reduce an SAE's reconstruction error for token $t$.
In a restricted \emph{context correction} experiment, we add a contribution from the feature activations $\tau$ tokens earlier to the reconstruction in \Cref{eq:standard-reconstruction}:
\begin{equation}
    \hat{\mathbf{x}}^{(t)}_{\mathrm{ctx}}
    =
    \hat{\mathbf{x}}^{(t)}_{\mathrm{base}}
    +
    \mathbf{W}_{\mathrm{dec}}
    \left(\mathbf{w}_\tau \odot \mathbf{f}^{(t-\tau)}\right).
    \label{eq:context-head}
\end{equation}
The weights $\mathbf{w}_\tau\in\mathbb{R}^m$ measure how much each past feature activation contributes to a later reconstruction along the same feature direction.
% Therefore, the weight vector $\mathbf{w}_\tau$ directly tests whether a feature's past activation predicts a later contribution along the same direction. 
For a given lag, we fit $\mathbf{w}_\tau$ for 10 epochs with an $L_2$ penalty and measure the fraction of reconstruction error that the context correction recovers on held-out data:\looseness=-1
\begin{equation}
    R^2_{\mathrm{improve}}
    =
    1 -
    \Big({
        \sum_t \lVert \mathbf{x}^{(t)}-\hat{\mathbf{x}}^{(t)}_{\mathrm{ctx}}\rVert_2^2
    }\Big) \Big/ \Big({
        \sum_t \lVert \mathbf{x}^{(t)}-\hat{\mathbf{x}}^{(t)}_{\mathrm{base}}\rVert_2^2
    }\Big).
    \label{eq:held-out-fraction-correction}
\end{equation}
% \begin{equation}
%     R^2_{\mathrm{improve}}
%     =
%     1 -
%     \frac{
%         \sum_t \lVert \mathbf{x}^{(t)}-\hat{\mathbf{x}}^{(t)}_{\mathrm{ctx}}\rVert_2^2
%     }{
%         \sum_t \lVert \mathbf{x}^{(t)}-\hat{\mathbf{x}}^{(t)}_{\mathrm{base}}\rVert_2^2
%     }.
%     \label{eq:held-out-fraction-correction}
% \end{equation}
For control, we use past activations from shuffled documents while keeping the same natural-document hidden states as targets. This isolates whether the reconstruction gain depends on the true preceding context.
% For the control, we refit the context correction weights using past activations from the shuffled control while keeping the natural document hidden states as targets. This tests whether the reconstruction gain requires the actual document history. 
We describe the training details in \Cref{app:context-correction-details}.

In \Cref{fig:main-observation}c, we observe that past activations from natural documents explain part of the reconstruction error left by the standard SAE, with gains decreasing as the lag grows. However, this effect is absent when training on shuffled control activations. Correspondingly, \Cref{fig:main-observation}d shows an example at lag $\tau=1$ where the fitted context-correction weights are predominantly positive, indicating that past feature activations continue to contribute along the same feature directions. Together, these results show that sparse features contain temporal information that is not explicitly modelled by the SAE.

We note that recent work \citep{priors_in_time} characterises temporal dependence at the \emph{hidden state level}, showing that correlations vary across a sequence and that past hidden states explain part of the current hidden state. In contrast, we study the \emph{feature level}, asking whether individual SAE features exhibit analogous persistence and whether their past activations help reconstruct later hidden states. 
\rebuttal{This resolution matters for interpretability: a hidden-state correlation aggregates all features, so two sequences can show the same temporal correlation even when different features produce it.}
We find substantial variation in persistence across features, motivating the explicit learning of a separate timescale for each feature in the next section. 
\rebuttal{We note that the feature-level view also leads to a much \emph{simpler} and interpretable model. Specifically, Persistent SAEs add one coefficient per feature ($16$k parameters), while the attention module of \citet{priors_in_time} (TFA) adds about $1$B parameters when using the same dictionary size. Crucially, even with the larger capacity, TFA reconstructs activations less accurately than Persistent SAEs when given an equivalent budget (see \Cref{app:tfa-comparison} for a detailed comparison).}
% do individual features capture this temporal structure. We find that their persistence helps reconstruction and varies widely across features, motivating the explicit learning of temporal persistence for each feature in the next section.

\section{Persistent Sparse Autoencoders}
\label{sec:method}

\begin{figure*}[t]
    \centering
    \includegraphics[width=\linewidth]{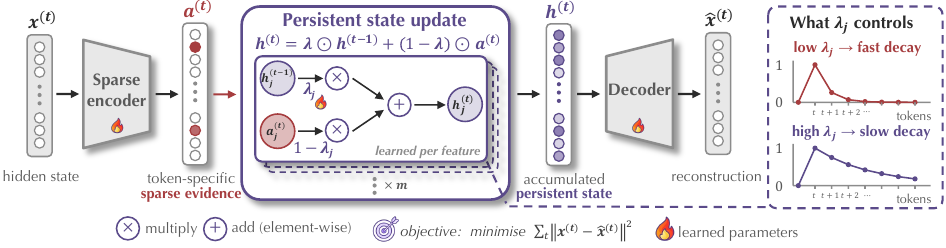}
    \caption{
    \textbf{Persistent Sparse Autoencoders.} Each feature integrates its token-wise sparse activation $\mathbf{a}^{(t)}$ into a persistent state $\mathbf{h}^{(t)}$ using its learned persistence coefficient $\lambda_j$. This allows different features to retain information over different timescales.
    }
    \label{fig:persistent-sae-architecture}
\end{figure*}

In this section, we introduce Persistent SAEs, which make the temporal structure observed in the previous section explicit by separating a token-wise sparse activation from a persistent \emph{state} (\Cref{fig:persistent-sae-architecture}).
%%%%%%%%%
%%%%%%%%%
As in standard BatchTopK encoders (see \Cref{app:batchtopk} for details), a Persistent SAE encoder produces token-wise sparse activation $\mathbf{a}^{(t)}\in\mathbb{R}_{\geq0}^m$, where $a^{(t)}_j>0$ means feature $j$ is active at token $t$. In contrast to standard SAEs, however, each feature integrates this activation into a \emph{persistent state} $\mathbf{h}^{(t)}\in\mathbb{R}_{\geq0}^m$ using a learned persistence coefficient $\lambda_j \in [0, 1]$:
\begin{equation}
    h^{(t)}_j
    =
    \lambda_j h^{(t-1)}_j
    +
    (1-\lambda_j)a^{(t)}_j.
    \label{eq:state-update}
\end{equation}
% This is the leaky-integrator form long used to model drifting context in human memory \citep{howard2002distributed} and, with a learned per-dimension decay, in modern sequence models \citep{ma2023mega,gu2024mamba}.
The decoder then reconstructs the hidden state from this persistent state: $\hat{\mathbf{x}}^{(t)} = \mathbf{W}_{\mathrm{dec}}\mathbf{h}^{(t)} + \mathbf{b}_{\mathrm{dec}}$.

Intuitively, a small $\lambda_j$ yields a \emph{short-timescale fast} feature that stays close to the current token, while a large $\lambda_j$ gives a \emph{long-timescale slow} feature that integrates over many tokens across the sequence. 
% (\Cref{sec:recon-quality} makes the slow and fast groups precise)
% we use \emph{fast} and \emph{slow} as shorthand for these two ends throughout.
% After $k$ tokens, a past activation is attenuated by $\lambda_j^k$, corresponding to an exponential timescale of $-1/\log \lambda_j$ tokens. 
The factor $(1-\lambda_j)$ prevents activations from accumulating excessively, keeping the state magnitude bounded.
During training, we initialise $\mathbf{h}^{(0)}=\mathbf{0}$ at the start of every input sequence. We parametrise $\lambda_j=\lambda_{\max}\sigma(l_j)$, where $l_j$ is an unconstrained learned parameter and $\lambda_{\max}=0.999$ keeps every feature's timescale finite. \Cref{app:persistence-param} provides further details about the persistence parametrisation.

The recurrence is diagonal in feature space: feature $j$ carries only its own activation forward, preserving feature-level attribution through its corresponding decoder direction. Sparsity is imposed only on the token-wise activation $\mathbf{a}^{(t)}$, so the accumulated state $\mathbf{h}^{(t)}$ is not directly sparsified. 
\Cref{app:sae-baselines} shows that it remains bounded in practice: the number of active features in the state stabilises at a small fraction of the dictionary during training.
% This construction also subsumes the traditional approach: setting every $\lambda_j$ to $0$ recovers a token-specific SAE.
Intuitively, the objective favours a larger $\lambda_j$ when retaining feature $j$ helps reconstruct later hidden states. The model therefore discovers each feature's timescale without the predefined temporal roles that previous work requires.

\section{Experiments}
\label{sec:experiments}

We evaluate the quality of the representations learnt by Persistent~SAEs and explore what appears at each learned timescale. We first test whether adding a persistent state preserves reconstruction quality and interpretable token-wise activations (\Cref{sec:recon-quality,sec:fast-features}). We then ask what information the accumulated states of slow features capture (\Cref{sec:slow-features}). Finally, we demonstrate their practical use in long-context monitoring through a prompt injection case study (\Cref{sec:prompt-injection}).

\myparagraph{Models and hyperparameters.}
For SAE training, we use the models and dataset from \Cref{sec:temporal-persistence}. We compare with BatchTopK \citep{batchtopk_sae} and Matryoshka \citep{matryoshka_sae} as standard SAE baselines, and Temporal SAE \citep{temporal_sparse_autoencoders} as the closest temporally-structured baseline. 
\rebuttal{We exclude TFA~\citep{priors_in_time} from our main evaluation because it differs fundamentally from the SAE baselines in how it reconstructs hidden states. Rather than reconstructing from a sparse activation, TFA reconstructs from the sum of a dense predicted code and a sparse novel code, making standard SAE metrics not directly comparable.
Nevertheless, we include a comparison on the quality of reconstruction of TFA compared to Persistent SAEs in \Cref{app:tfa-comparison}, where we show that Persistent SAEs attain better reconstruction scores than TFA.}
All SAEs use $m\!=\!16{,}384$ features and a BatchTopK budget of $k\!=\!20$. Following \citep{temporal_sparse_autoencoders}, Matryoshka uses a two-level nested dictionary with 20\% of features in the smaller inner dictionary, while Temporal SAE assigns 20\% of features to a predefined high-level group and the remaining 80\% to a low-level group.
Persistent SAEs are trained on contiguous sequences of 256 tokens, which determines how many tokens the persistent state can propagate across during each training sample.
For a fair comparison, we match the total training token budget across SAE families.
\Cref{app:training-details} gives the full training configuration.
% We show that Persistent SAE training stabilises without continual growth in the persistent state.

\subsection{Reconstruction Quality and Learned Timescales}
\label{sec:recon-quality}

\begin{figure*}[!t]
    \centering
    \includegraphics[width=\linewidth]{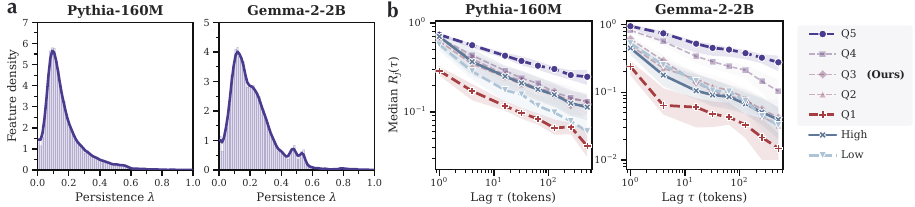}
    \caption{
        \textbf{Learned persistence coefficients align with empirical persistence.} \textbf{(a)}~Persistence coefficients $\lambda_j$ span a broad range, with many short-timescale features and fewer long-timescale features. \textbf{(b)}~We divide Persistent SAE features into quintiles by their learned $\lambda_j$. Higher-$\lambda_j$ quintiles retain greater median empirical persistence $R_j(\tau)$ across different lags. 
        % Lines show medians over equally sized feature samples, and shaded regions show 95\% feature-bootstrap intervals.
}
\label{fig:timescale-co-activation}
\end{figure*}

We first examine how the learned persistence coefficients are distributed across features. \Cref{fig:timescale-co-activation}a shows that most features are short-timescale and fast, with a small tail of long-timescale, highly persistent slow features.
Moreover, we observe that similar distributions appear across different initialisations (\Cref{app:persistence-init-ablation}),
% and training sequence lengths (\Cref{app:context-length-ablation}).
training sequence lengths (\Cref{app:context-length-ablation}), and model scales (\Cref{app:model-size-persistence}).
\rebuttal{We also compare against simpler post-hoc baselines that add an EMA with fixed or learned $\lambda_j$ to a pretrained standard SAE, showing that the Persistent SAE's advantage comes from jointly learning features and their timescales (\Cref{app:posthoc-ema}).}
% To compare the small slow tail with the broader population of fast features in subsequent analyses,
To facilitate analysis and more easily compare our features with those discovered by competing baselines (e.g., high- and low-level features from Temporal SAEs), we split features into two groups considering the observed skewness in their distributions. Specifically, 
% To compare this slow tail with the rest of the dictionary,
we define ``slow'' features as the highest $1\%$ by $\lambda_j$ and ``fast'' features as the lowest $50\%$.
% These cutoffs are a convenience for baseline comparison:
However, we emphasise that in practice $\lambda_j$ varies continuously, and show in \Cref{app:gemma2b-semantic} that in-group semantic cohesiveness is preserved across thresholds.
% in \Cref{fig:semantic-slow-feature-sweep} in \Cref{app:tsne-details} we show that the results change gradually as the cutoff moves.
% The learned persistence distribution is also robust to two potential sources of variation: it is similar across initialisations (\Cref{app:persistence-init-ablation}) and stable across training sequence lengths (\Cref{app:context-length-ablation}).
% The temporal structure discovered by Persistent SAEs is therefore robust to both choices.

We then evaluate representation quality using: (1)~fraction of variance explained (FVE); (2)~cosine similarity between the hidden state and their reconstruction; (3)~fraction of active features; and (4)~how smoothly features change over time, using the temporal smoothness metric from \citet{temporal_sparse_autoencoders}, which we define in \Cref{app:smoothness-details}, 
\rebuttal{and (5) the average number of active features per token ($L_0$).}
\Cref{tab:core-performance} shows that Persistent~SAEs remain competitive with the strongest baselines in reconstruction quality.
\rebuttal{Because the decoder reads the persistent state, which has more non-zero entries than the token-wise activation, we also restrict it to the same $20$ entries per token as the baselines and find that its reconstruction stays within the range of the standard SAEs (\Cref{app:read-budget}).} 
Across both models, \rebuttal{the state of} slow features changes more smoothly over time than fast features and baseline SAEs. 
% \rebuttal{On the token-wise activation, slow features are less smooth than the baselines, so this smoothness comes from integrating activations over time (\Cref{app:smoothness-details}, \Cref{tab:smoothness-state-activation}).} 
This suggests that Persistent SAEs preserve fast features for token-specific detail while concentrating temporal continuity in a small slow subset, without preassigning temporal roles. 
% \rebuttal{This smoothness comes from integrating token-wise activations over time: on the token-wise activation $\mathbf{a}^{(t)}$ itself, slow features score $0.21$ on Pythia and $0.26$ on Gemma, close to or above fast features ($0.22$ and $0.20$; \Cref{app:smoothness-details}).}
% Moreover, our results show that high-$\lambda_j$ (slow) features have smoothness $0.05$ on Pythia and $0.04$ on Gemma, whereas low-$\lambda_j$ (fast) features score $0.20$ on both models. This suggests that, as desired, Persistent SAE preserves fast features for token-specific detail while concentrating temporal continuity in a smaller, slower subset, all without requiring a user to predefine which features should decay slowly and which ones should not.

\begin{table*}[!t]
\centering
\footnotesize
\setlength{\tabcolsep}{1.2pt}
\renewcommand{\arraystretch}{1.12}
\caption{
\textbf{SAE representation quality}. \rebuttal{\emph{Slow}/\emph{Fast} denote the highest $1\%$/lowest $50\%$ of Persistent SAE features ranked by $\lambda_j$. For the Temporal SAE, we use its predefined high-/low-level feature partition (\Cref{app:sae-baselines}) as the slow/fast features.}
% Matryoshka SAE has no comparable per-feature split and is reported only under \emph{Full}. 
% Error bars are approximate 95\% confidence half-widths from evaluation-sample variability. 
\rebuttal{$L_0$ is the average number of non-zero activations per token.}
\rebuttal{Smoothness is evaluated on $\mathbf{h}^{(t)}$ for Persistent SAE.}
Best/second-best results are bold/underlined. Persistent SAEs remain competitive while separating smooth, slow features from locally interpretable fast features.
}
\resizebox{\textwidth}{!}{%
\begin{tabular}{llccccccccccc}
\toprule
&
&
\multirow{2}{*}{FVE ($\uparrow$)}
&
\multirow{2}{*}{\shortstack{Cosine\\Sim ($\uparrow$)}}
&
\multirow{2}{*}{\shortstack{Fraction\\Alive ($\uparrow$)}}
&
\multicolumn{3}{c}{Smoothness ($\downarrow$)}
&
\multicolumn{3}{c}{Autointerp ($\uparrow$)}
&
\multirow{2}{*}{\rebuttal{$L_0$}}
\\[-1pt]
\cmidrule(lr){6-8}
\cmidrule(lr){9-11}
&
&
&
&
&
Full
&
\rebuttal{Slow}
&
\rebuttal{Fast}
&
Full
&
\rebuttal{Slow}
&
\rebuttal{Fast}
&
\\
\midrule

\multirow{4}{*}{
    \rotatebox[origin=c]{90}{
        Pythia
    }
}
& BatchTopK
& $\mathbf{0.94{\scriptscriptstyle\pm.01}}$
& $\mathbf{0.95{\scriptscriptstyle\pm.01}}$
& $\underline{0.91{\scriptscriptstyle\pm.01}}$
& $0.15{\scriptscriptstyle\pm.01}$
& --
& --
& $\mathbf{0.91{\scriptscriptstyle\pm.02}}$
& --
& --
& \rebuttal{$19.0{\scriptscriptstyle\pm.2}$}
\\

& Matryoshka
& $0.92{\scriptscriptstyle\pm.01}$
& $\underline{0.94{\scriptscriptstyle\pm.01}}$
& $\mathbf{0.96{\scriptscriptstyle\pm.01}}$
& $0.15{\scriptscriptstyle\pm.01}$
& --
& --
& $\underline{0.87{\scriptscriptstyle\pm.02}}$
& --
& --
& \rebuttal{$19.3{\scriptscriptstyle\pm.1}$}
\\

& Temporal
& $\underline{0.93{\scriptscriptstyle\pm.02}}$
& $\mathbf{0.95{\scriptscriptstyle\pm.01}}$
& $0.79{\scriptscriptstyle\pm.01}$
& $\underline{0.13{\scriptscriptstyle\pm.01}}$
& $\underline{0.18{\scriptscriptstyle\pm.01}}$
& $\mathbf{0.11{\scriptscriptstyle\pm.01}}$
& $\underline{0.87{\scriptscriptstyle\pm.02}}$
& $\mathbf{0.87{\scriptscriptstyle\pm.01}}$
& $\underline{0.89{\scriptscriptstyle\pm.02}}$
& \rebuttal{$19.1{\scriptscriptstyle\pm.2}$}
\\

& \cellcolor{gray!10}Persistent (Ours)
& \cellcolor{gray!10}$\underline{0.93{\scriptscriptstyle\pm.01}}$
& \cellcolor{gray!10}$\mathbf{0.95{\scriptscriptstyle\pm.01}}$
& \cellcolor{gray!10}$\mathbf{0.96{\scriptscriptstyle\pm.01}}$
& \cellcolor{gray!10}$\mathbf{0.10{\scriptscriptstyle\pm.01}}$
& \cellcolor{gray!10}$\mathbf{0.05{\scriptscriptstyle\pm.01}}$
& \cellcolor{gray!10}$\underline{0.20{\scriptscriptstyle\pm.01}}$
& \cellcolor{gray!10}$0.86{\scriptscriptstyle\pm.01}$
& \cellcolor{gray!10}$\underline{0.84{\scriptscriptstyle\pm.02}}$
& \cellcolor{gray!10}$\mathbf{0.93{\scriptscriptstyle\pm.01}}$
& \cellcolor{gray!10}\rebuttal{$22.3{\scriptscriptstyle\pm.3}$}
\\

\midrule

\multirow{4}{*}{
    \rotatebox[origin=c]{90}{
        Gemma
    }
}
& BatchTopK
& \rebuttal{$\mathbf{0.73{\scriptscriptstyle\pm.01}}$}
& \rebuttal{$\mathbf{0.89{\scriptscriptstyle\pm.01}}$}
& $\mathbf{0.92{\scriptscriptstyle\pm.01}}$
& \rebuttal{$0.14{\scriptscriptstyle\pm.01}$}
& --
& --
& $\underline{0.87{\scriptscriptstyle\pm.03}}$
& --
& --
& \rebuttal{$19.7{\scriptscriptstyle\pm.2}$}
\\

& Matryoshka
& \rebuttal{$\underline{0.72{\scriptscriptstyle\pm.01}}$}
& \rebuttal{$\mathbf{0.89{\scriptscriptstyle\pm.01}}$}
& $0.87{\scriptscriptstyle\pm.01}$
& \rebuttal{$0.15{\scriptscriptstyle\pm.01}$}
& --
& --
& $\mathbf{0.89{\scriptscriptstyle\pm.02}}$
& --
& --
& \rebuttal{$19.9{\scriptscriptstyle\pm.2}$}
\\

& Temporal
& \rebuttal{$0.70{\scriptscriptstyle\pm.01}$}
& \rebuttal{$\underline{0.88{\scriptscriptstyle\pm.01}}$}
& $0.75{\scriptscriptstyle\pm.01}$
& \rebuttal{$\underline{0.13{\scriptscriptstyle\pm.01}}$}
& \rebuttal{$\underline{0.10{\scriptscriptstyle\pm.01}}$}
& \rebuttal{$\mathbf{0.15{\scriptscriptstyle\pm.01}}$}
& $0.85{\scriptscriptstyle\pm.02}$
& $\mathbf{0.85{\scriptscriptstyle\pm.02}}$
& $\underline{0.86{\scriptscriptstyle\pm.02}}$
& \rebuttal{$19.9{\scriptscriptstyle\pm.2}$}
\\

& \cellcolor{gray!10}Persistent (Ours)
& \cellcolor{gray!10}\rebuttal{$\mathbf{0.73{\scriptscriptstyle\pm.01}}$}
& \cellcolor{gray!10}\rebuttal{$\mathbf{0.89{\scriptscriptstyle\pm.01}}$}
& \cellcolor{gray!10}$\underline{0.91{\scriptscriptstyle\pm.01}}$
& \cellcolor{gray!10}\rebuttal{$\mathbf{0.12{\scriptscriptstyle\pm.01}}$}
& \cellcolor{gray!10}\rebuttal{$\mathbf{0.03{\scriptscriptstyle\pm.01}}$}
& \cellcolor{gray!10}\rebuttal{$\underline{0.18{\scriptscriptstyle\pm.01}}$}
& \cellcolor{gray!10}$0.85{\scriptscriptstyle\pm.01}$
& \cellcolor{gray!10}$\underline{0.79{\scriptscriptstyle\pm.01}}$
& \cellcolor{gray!10}$\mathbf{0.89{\scriptscriptstyle\pm.01}}$
& \cellcolor{gray!10}\rebuttal{$22.6{\scriptscriptstyle\pm.3}$}
\\

\bottomrule
\end{tabular}%
}

\label{tab:core-performance}
\end{table*}

Moreover, we test whether the learned coefficients correspond to the empirical persistence observed in \Cref{sec:temporal-persistence}. We group features into quintiles by $\lambda_j$ and measure their median $R_j(\tau)$ across lags. \Cref{fig:timescale-co-activation}b shows a consistent ordering across both models: higher-$\lambda_j$ groups decay more slowly and remain more persistent over longer lags. By contrast, the Temporal SAE's predefined high-level and low-level groups are not clearly separated by empirical persistence. This alignment shows that the coefficients learnt through reconstruction capture meaningful differences in feature timescales.

% A natural question is whether multiple features could share the same persistence coefficient. We test this by varying the number of features that share each coefficient. As fewer features are forced to share a coefficient, the model recovers a progressively wider range of empirical timescales (\Cref{app:persistence-granularity}).
% This suggests that different features require different persistence rates, supporting our choice to learn a separate coefficient for each feature.
% We further train Persistent SAEs on ten models from three families, ranging from 160M to 12B parameters (\Cref{app:model-size-persistence}). Every model learns a broad spectrum of timescales, while larger models shift this spectrum towards more persistent features, suggesting that they allocate more features to information that persists over longer contexts.

\subsection{Fast Features Are Locally Interpretable}
\label{sec:fast-features}

We next ask how learned timescales relate to feature interpretability. We use the automated interpretability pipeline of \citet{bills2023language}, in which a judge model explains a feature from its highest-activation contexts and then scores how well that explanation predicts activation. For the Persistent~SAE, we evaluate the token-wise sparse activation $\mathbf{a}^{(t)}$, asking whether it remains understandable before it is used to update the persistent state (see \Cref{app:autointerp-details} for details on autointerpretation).\looseness=-1

Our experiments, summarised in \Cref{tab:core-performance} (right), suggest that, across both models, fast features are easier to explain locally than slow features. 
Thus, adding recurrence preserves locally interpretable activations for fast features, which match or exceed the best baseline score on both models, while interpretability varies systematically with the learned timescale.
% More importantly, we observe that fast features match or exceed the best baseline score on both models.
Qualitative examples (Tables~\ref{tab:autointerp-qualitative-pythia160m} and~\ref{tab:autointerp-qualitative-gemma2b} in \Cref{app:autointerp-details}) suggest that fast features often respond to compact lexical or grammatical patterns, such as letters or substrings. 
% Their evidence is visible at the current token and can be precisely captured by a short description. 
% Their activation is concentrated at the current token, making these features easy to characterise from short local contexts.
In contrast, slow features more often reflect broader subjects whose activation is distributed across the sequence.
This exposes a mismatch between persistent representations and a token-specific feature evaluation:
distinct features may receive similar descriptions \citep{mccann2026descriptive}, and explanations may not generalise to related inputs \citep{tian2025sensitivity}. Hence, in \Cref{sec:slow-features} we study the slow features as a group and test what information their state carries.

\myparagraph{Where features are activated.}
% The preceding analysis asks how well individual feature activations can be explained at the tokens where they occur. 
Before examining the states of slow features, we ask which tokens activate features with different timescales. We split features into five equal-sized quintiles by $\lambda_j$ and assign each token to one of six categories: punctuation, conjunctions, function words (\emph{Func.}), verbs, adjectives or adverbs, and nouns (see \Cref{app:token-type-details} for details). For each quintile $q$ and category $c$, we compute the relative activation frequency
\begin{equation}
    U_{q,c} = {\Pr\!\left(T^{(t)}=c \mid a^{(t)}_j>0,\;j\in Q_q\right)} \big/ {\Pr\!\left(T^{(t)}=c\right)},
\label{eq:activation-relative-frequency}
\end{equation}
\begin{wrapfigure}{r}{0.4\textwidth}
    \centering
    \includegraphics[width=\linewidth]{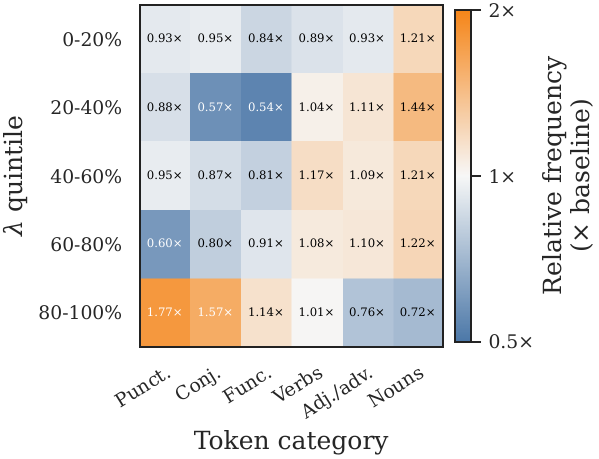}
    \caption{\textbf{Where features are activated.} Each cell shows how often features in a $\lambda$-quintile activate at a token category relative to that category's corpus-level frequency.  \textcolor{orange!70!yellow}{\textbf{Orange}}/\textcolor{blue!60!cyan}{\textbf{blue}} indicate frequencies above/below baseline. Faster features favour nouns; slower features favour punctuation and conjunctions.}
    \label{fig:activation-token-types}
    \vspace*{-6mm}
\end{wrapfigure}
where $T^{(t)}$ is the category of token $t$ and $Q_q$ is the set of features in quintile $q$. A value above one means that features in quintile $q$ activate at category $c$ more often than expected from how common $c$ is in the corpus.\looseness=-1

\Cref{fig:activation-token-types} shows the difference between the two ends of the persistence distribution. Faster features are active more often at nouns. This aligns with their stronger local interpretability in \Cref{tab:core-performance} as nouns carry clear lexical content at the current token. Slower features, in contrast, are active more often at punctuation and conjunctions. 
Recent analyses of LLMs likewise find that function-like tokens can carry substantial contextual information and structural markers integrate information across sentences \citep{razzhigaev-etal-2025-llm, liu-ding-2025-information}.
Thus, faster features are relatively more active at lexical content, whereas slower features are preferentially active at positions associated with organising and integrating context. 
% We next examine what these accumulated states represent.
However, where a feature activates does not tell us what its state encodes. The automated explanations suggest that slow features capture broad topics. Therefore, we test whether their states track the subject of the sequence in the next section.\looseness=-1 

\subsection{States of Slow Features Track Subject Information}
\label{sec:slow-features}

\begin{figure*}[!t]
    \centering
    \includegraphics[width=\linewidth]{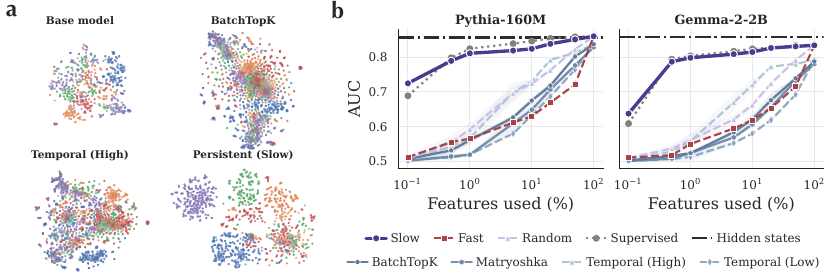}
    \caption{
    \textbf{The state of the slow features tracks subject information.}
    \textbf{(a)}~t-SNE of token-level MMLU representations, coloured by subject. The state of the Persistent SAE's slow features (high~$\lambda_j$) reveals clearer subject structure than baseline SAEs or the raw hidden states.
    \textbf{(b)}~Subject-probing AUC on spliced MMLU sequences. Selecting a fraction of features in decreasing order of $\lambda_j$ recovers most of the subject signal, approaching supervised feature selection.
    }
    \label{fig:slow-features}
\end{figure*}

\myparagraph{Slow features form subject-specific clusters.}
To test whether the accumulated states of slow features retain higher-level contextual information, we visualise the slow-feature states on MMLU \citep{MMLU} examples from five subjects. For each sample, we collect the representations of its final $20$ tokens and visualise them using t-SNE (see \Cref{app:tsne-details} for details). We use slow features (the highest $1\%$ by $\lambda_j$) for visualisation. \Cref{fig:slow-features}a shows that BatchTopK features heavily mix subjects, whereas the Temporal SAE's high-level group shows limited subject separation. In contrast, the state of Persistent SAE's slow features forms coherent subject regions with a clearer visual separation than even the raw hidden states. 
Importantly, this structure is learned \textit{without} subject labels.\looseness=-1
%%%%%%%%%%%%%%%%%%%%%%%%%%%%%%%

\myparagraph{Learned persistence provides unsupervised selection of subject-informative features.}
We test this observation quantitatively on spliced MMLU sequences, created by concatenating labelled segments from different subjects so that the subject changes at segment boundaries. At each token, a linear probe predicts the current subject using varying fractions of the feature dictionary for training the probe. 
We compare three rankings of Persistent SAE features: decreasing $\lambda_j$ (slowest first), increasing $\lambda_j$ (fastest first), and random ordering.
For baseline SAEs which do not provide a criterion for feature selection, we select features in random order.
For Temporal SAE, we rank either the high-level group before the low-level group, or vice versa, with random ordering within each group.
We also include a supervised ranking that uses subject labels to favour features that activate more strongly on each subject than on all other subjects. This serves as a supervised oracle baseline (see \Cref{app:probing-details} for details).\looseness-1

As seen in \Cref{fig:slow-features}b, selecting features by decreasing persistence is nearly as efficient as supervised selection, while the \mbox{increasing-$\lambda_j$}, random, and baseline rankings retain substantially less subject information at the same budget.
Interestingly, the \mbox{increasing-$\lambda_j$} ranking prioritises the more locally interpretable fast features but is often less efficient than baseline SAEs, suggesting that different feature persistences support a division of labour: faster features capture local, interpretable signals, while slower features retain higher-level contextual information.
\Cref{app:token-similarity} extends this result with a qualitative analysis to show how the state of the slow features behaves across subject boundaries. We find that tokens from the same subject have similar states, while tokens from different subjects do not.
\rebuttal{Moreover, in \Cref{app:group-interaction}, we show that features of the Persistent SAE form topic groups that stay active together while the subject continues, even though each feature is designed to persist on its own.}
Thus, persistence provides an unsupervised criterion for selecting a compact state from a much wider sparse dictionary that tracks the current subject of a sequence.

\subsection{Case Study: Persistent Prompt Injection Monitoring}
\label{sec:prompt-injection}

The preceding experiments suggest that fast features detect interpretable patterns at the current token, while slow features integrate token-wise activations into a compact persistent state. Indirect prompt injection \citep{agentdojo} provides a practical test of this hypothesis. In this setting, an agent may encounter a malicious instruction in an email, webpage, or tool response, continue reasoning for hundreds of tokens, and act later \citep{camel_prompt_injection}. A useful agentic monitor must therefore do more than detect the injected text when it appears and should preserve a security-relevant signal until the agent acts \citep{injecagent,ye2026prompt}.

\myparagraph{Setting.}
We use AgentDojo trajectories from the banking, travel, and workspace domains \citep{agentdojo}, generated by Llama-3.3-70B-Instruct \citep{Llama3} (see \Cref{app:agentdojo-details} for details). Each trajectory contains a request, an untrusted span with a prompt injection or benign content, and the agent's subsequent tool calls. 
%To prevent the monitor from relying on malicious actions to identify an attack, we consider only failed attacks, where an injection is present but the agent does not complete the attack's action.
To prevent detection based on malicious actions, we use only failed attacks: injections that do not lead the agent to complete the attacker's intended action.\looseness-1
% This prevents the monitor from relying on an overtly malicious action to identify an attack.

\begin{figure*}[!t]
    \centering
    \includegraphics[width=\linewidth]{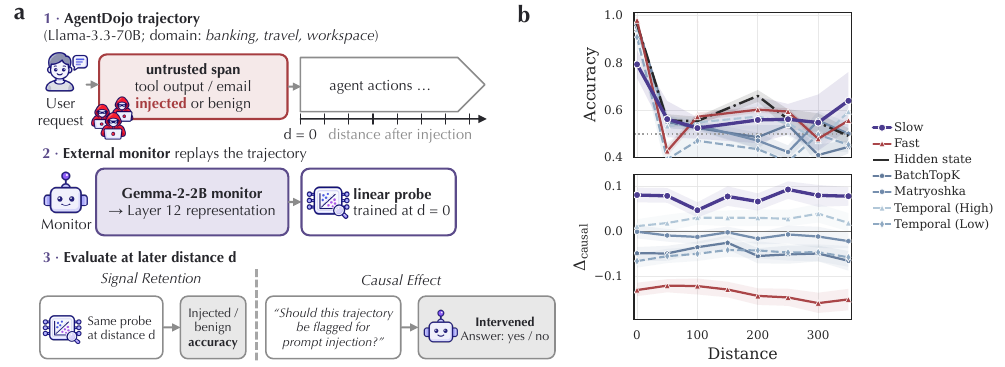}
    \caption{
        \textbf{Persistent monitoring of prompt injection.} \textbf{(a)} A probe trained at the end of an untrusted span is evaluated later in the trajectory. \textbf{(b)} Slow-feature states retain more predictive information (top), while their decoder directions exert the strongest causal influence on the monitor's judgment (bottom).\looseness-1
    }
    \label{fig:prompt-injection-monitoring}
\end{figure*}

We use Gemma-2-2B as an external monitor: we replay each trajectory through the monitor model and record the layer-12 residual stream. We train a linear probe at the final token of the untrusted span ($d=0$) and evaluate the same probe up to $350$ tokens later without retraining (\Cref{fig:prompt-injection-monitoring}a). 
We compare the hidden state and baseline SAEs with the slow- and fast-feature Persistent SAE states as defined in \Cref{sec:recon-quality}. 
This evaluation measures how long each representation preserves the injection signal (see \Cref{app:prompt-injection-details} for evaluation details).

\myparagraph{Signal retention: fast features detect local injection signals while slow features retain signals over long contexts.}
At $d=0$, fast features and standard SAE representations provide the strongest signals for detecting an injection (\Cref{fig:prompt-injection-monitoring}b, top). As the evaluation moves further beyond the untrusted span, their accuracy falls towards chance. By contrast, slow features retain a clearer injection signal at longer distances. This result again confirms that fast and slow features play complementary roles: fast features detect injected text when it appears, while the state of slow features preserves the signal through the subsequent context.
\rebuttal{We hypothesise that this retention arises from both the decayed state carried from earlier tokens and continued activation of the same slow features. These two sources are complementary: later activations can reflect earlier context that the LLM's hidden states already carry, and the Persistent SAE then re-activates the same slow feature instead of representing this context with a new feature.}

% At $d=0$, the Persistent SAE's token-wise activations and the token-specific SAE baselines reach 94.0--96.7\% accuracy (\Cref{fig:prompt-injection-monitoring} (b), top). Thus, the token-wise activation (a proxy for fast features in Persistent SAEs) is as good a signal for detecting an injection as locally interpretable detectors when the representations are extracted from near the boundary of the injection text span. As the evaluation moves beyond the untrusted span, it and the token-specific representations fall towards chance. In contrast, the slow features remain informative across longer distances. Fast and slow features therefore play complementary roles: fast features detect the injection locally, while slow features carry its signal through the subsequent context.

\myparagraph{Causal effect: slow features influence the monitor's judgement.}
Probing shows that prompt-injection information can be decoded from the representations, but does not establish whether this information affects the monitor's decision. 
We therefore perform a causal intervention experiment.
At each distance $d$, we append a query asking the monitor whether a prompt injection occurred and intervene on the hidden state at the final query token. 
Building on prior work that selects SAE features for steering \citep{sae_right_features_steering, he-etal-2025-sae}, for each representation $r$, we construct a task-relevant steering direction $\mathbf{v}_r$ from features identified by the probe (see \Cref{app:prompt-injection-details} for details).
% Building on prior work that selects SAE features for steering and uses linear classifiers to identify task-relevant sparse subspaces \citep{sae_right_features_steering, he-etal-2025-sae}, we construct a steering vector $\mathbf{v}_r$ for each representation $r$ by weighting its top probe-selected feature directions by their probe coefficients. 
We apply interventions in both positive and negative directions. Let $s(\mathbf{x})=\ell_{\mathrm{yes}}(\mathbf{x})-\ell_{\mathrm{no}}(\mathbf{x})$ denote the monitor's logit margin. We measure the intervention effect as
\begin{equation}
    \begin{aligned}
        \Delta_{\mathrm{causal}}(r,d)
        = \frac{1}{2}\mathbb{E}_{i}\Big[
        s\big(\mathbf{x}_i^{(d)}+\alpha_i^{(d)}\mathbf{v}_r\big)
        -s\big(\mathbf{x}_i^{(d)}-\alpha_i^{(d)}\mathbf{v}_r\big)
        \Big].
    \end{aligned}
    \label{eq:causal-intervention-monitor}
\end{equation}
where $\alpha_i^{(d)}$ scales the intervention to $10\%$ of the hidden-state norm. A positive value means that moving along the direction makes the monitor likely to answer \textit{yes}.

Slow features produce the largest and most consistently positive intervention effect across all distances (\Cref{fig:prompt-injection-monitoring}b, bottom).
In contrast, directions derived from baseline SAE features produce small or negative effects.
These results complement the probing experiment. Fast features provide strong local evidence of an injection, whereas slow features preserve information that remains both decodable and causally relevant to the monitor's later decision. Persistent SAE therefore captures both short-lived local evidence and longer-lived state that can influence downstream model decisions, offering a sparse representation that may support more reliable monitoring and control of LLMs.

\section{Discussion and Conclusion}
\label{sec:discussion-conclusion}

\myparagraph{Discussion and Limitations.}
A natural question is whether multiple features could share a persistence coefficient. We test this by varying how many features are grouped to share a coefficient and find that, as groups get smaller, we recover a progressively wider range of empirical timescales (\Cref{app:persistence-granularity}).
This suggests that different features require different persistence rates, supporting our modelling choice.
% to learn a separate coefficient for each feature. 
We assign persistence coefficients through a diagonal recurrence, and therefore do not directly model cross-feature interactions.
\rebuttal{However, we show in \Cref{app:group-interaction} that features of the Persistent SAE already interact meaningfully, forming topic groups that follow the subject.}
% \rebuttal{However, features of the Persistent SAE already form groups that activate together within a few tokens and correspond to topics (\Cref{app:cross-feature}).}
\rebuttal{Although each $\lambda_j$ is fixed after training, in \Cref{app:token-similarity} we show that the state still adapts to context: when a feature stops receiving new activations, its state decays while features for the new context become active.}
\rebuttal{We also present an initial exploration of adding explicit cross-feature interaction in \Cref{app:cross-feature-correction}.}
Finally, we note that, in our prompt-injection case study, we use an external replay monitor of agent trajectories; therefore, we do not establish that persistent states have a causal effect on the original agent's actions. Future work could explore richer recurrent dynamics and test whether persistent feature interventions affect end-to-end agent behaviour.

\myparagraph{Conclusion.}
We introduced Persistent SAEs, which learn from reconstruction alone how long individual sparse features should retain information across tokens. The learned per-feature persistence coefficients span a broad range of timescales: short-timescale features remain locally interpretable, while long-timescale features accumulate contextual information, tracking subject identity and preserving monitoring signals over long contexts. As such, Persistent SAEs provide an unsupervised way to organise sparse representations by timescale and offer a potential direction for interpreting and monitoring information maintained by LLMs over long contexts.

\ifneurips\else
% \newpage
% \pagebreak
\subsection*{AI use statement}

% (This section is \textbf{required} and does not count toward the page limit.)

% In this work, we used generative AI tools for [tasks with required disclosure]. We have not used generative AI tools for [other tasks with required disclosure], and [the rest of the required disclosure tasks] are not applicable to this work. Additionally, we used generative AI tools for [tasks with recommended disclosure]. We have reviewed all AI-assisted work. [Elaborate. For example, ``we checked LLM-generated research ideas for potential plagiarism through a manual literature survey'', ``LLM-generated code was verified and tested for correctness by 2 authors'', etc.]. We take responsibility for the final content of this work, including text, claims or artefacts produced with the aid of generative AI.

% See the ICLR 2027 AI Policy for Authors for more details. This statement should not be more than 1 page.
In this work, we used generative AI tools as part of our experimental evaluation: \texttt{gpt-4o-mini} served as the judge model in the automated interpretability pipeline (Appendix~\ref{app:autointerp-details}), where it wrote short natural-language explanations of SAE features from their highest-activation contexts and then used these explanations to predict feature activations on held-out contexts. The judge model's outputs were used only to compute interpretability scores.
We also used generative AI tools to assist with coding. All AI-generated code was reviewed by the authors, verified against the intended implementation, and tested for correctness.
% We have not used generative AI tools for research ideation, and the remaining disclosure tasks are not applicable to this work.
Additionally, we used generative AI tools to correct grammatical errors in the manuscript. We have reviewed all AI-assisted work. All grammar corrections were checked by the authors, and no content, claims or results were generated by these tools. We take responsibility for the final content of this work, including text, claims or artefacts produced with the aid of generative AI.\looseness=-1

\subsection*{Reproducibility statement}
We have taken several steps to make our results reproducible. The Persistent SAE architecture and training objective are fully specified in \Cref{sec:method}, and \Cref{app:persistence-param} details the parameterisation and initialisation of the persistence coefficients. \Cref{app:training-details} gives the full training configuration for every SAE family, including the sparsity budget and token budget. All language models whose representations we analyse and all datasets we use are publicly available, and \Cref{app:dataset-details} describes the data processing steps for the Pile, MMLU, and AgentDojo data, including the construction of the shuffled control sequences. Appendices~\ref{app:persistence-derivation}--\ref{app:prompt-injection-details} give the evaluation procedures, and \Cref{app:ablation-studies} reports the sensitivity of our results to hyperparameters and model size.

% \subsection*{Acknowledgements}

% Here you need to acknowledge your funder and Mateo his. None for me. For now it can be commented out until final version. 

\fi

\ifneurips
  \bibliographystyle{unsrtnat}
\else
  \bibliographystyle{iclr2027_conference}
\fi
\ifneurips
\begin{ack}
H.L. acknowledges support from the Cambridge Trust. M.E.Z. acknowledges support from Trinity College, Oxford, as part of a Junior Research Fellowship (JRF).
\end{ack}
\fi

\bibliography{custom}

%%%%%%%%%%%%%%%%%%%%%%%%%%%%%%%%%%%%%%%%%%%%%%%%%%%%%
%%%%%%%%%%%%%%%%%%%%%%%%%%%%%%%%%%%%%%%%%%%%%%%%%%%%%
%%%%%%%%%%%%%%%%%%%%%%%%%%%%%%%%%%%%%%%%%%%%%%%%%%%%%
%%%%%%%%%%%%%%%%%%%%%%%%%%%%%%%%%%%%%%%%%%%%%%%%%%%%%
%%%%%%%%%%%%%%%%%%%%%%%%%%%%%%%%%%%%%%%%%%%%%%%%%%%%%
%%%%%%%%%%%%%%%%%%%%%%%%%%%%%%%%%%%%%%%%%%%%%%%%%%%%%
%%%%%%%%%%%%%%%%%%%%%%%%%%%%%%%%%%%%%%%%%%%%%%%%%%%%%
%%%%%%%%%%%%%%%%%%%%%%%%%%%%%%%%%%%%%%%%%%%%%%%%%%%%%
%%%%%%%%%%%%%%%%%%%%%%%%%%%%%%%%%%%%%%%%%%%%%%%%%%%%%
%%%%%%%%%%%%%%%%%%%%%%%%%%%%%%%%%%%%%%%%%%%%%%%%%%%%%
%%%%%%%%%%%%%%%%%%%%%%%%%%%%%%%%%%%%%%%%%%%%%%%%%%%%%
%%%%%%%%%%%%%%%%%%%%%%%%%%%%%%%%%%%%%%%%%%%%%%%%%%%%%
%%%%%%%%%%%%%%%%%%%%%%%%%%%%%%%%%%%%%%%%%%%%%%%%%%%%%
\newpage
\pagebreak

\appendix

\section{Notation}
\label{app:notation}

\Cref{tab:notation} lists the symbols used in the main text and appendices, in order of first appearance, as a reference for the terminology introduced in Sections~\ref{sec:temporal-persistence}--\ref{sec:prompt-injection}. 
% We write vectors and matrices in bold, index token positions with a parenthesised superscript, and index coordinates within a vector with a subscript.

\begin{table}[!h]
\centering
\small
\caption{Notation used throughout the paper, grouped by the section in which each symbol is introduced. Bold symbols denote vectors and matrices, a parenthesised superscript denotes a token position, and a subscript denotes a coordinate within a vector.}
\label{tab:notation}
\begin{tabular}{@{}ll@{}}
\toprule
\textbf{Symbol} & \textbf{Meaning} \\
\midrule
\multicolumn{2}{@{}l}{\textit{Hidden states and standard SAE (\Cref{sec:temporal-persistence}, \ref{sec:method})}} \\
$\mathbf{x}^{(t)} \in \mathbb{R}^d$ & Residual-stream hidden state at token position $t$ \\
$T$ & Number of tokens in a sequence \\
$d$ & Dimensionality of the hidden state \\
$m$ & Number of dictionary features (SAE width) \\
$\mathbf{W}_{\mathrm{enc}}, \mathbf{b}_{\mathrm{enc}}$ & Standard SAE encoder weight matrix and bias \\
$\mathbf{W}_{\mathrm{dec}}, \mathbf{b}_{\mathrm{dec}}$ & Standard SAE decoder weight matrix and bias \\
$\mathbf{d}_j$ & Unit-normalised decoder direction for feature $j$ \\
$\mathbf{f}^{(t)} \in \mathbb{R}_{\ge 0}^m$ & Sparse activation of a standard SAE at token $t$ (Background and Related Work) \\
$f^{(t)}_j$ & Sparse activation of feature $j$ at token $t$ \\
$k$ & BatchTopK sparsity level (\Cref{app:batchtopk}) \\
$B$ & Batch size used by BatchTopK selection (\Cref{app:batchtopk}) \\
\midrule
\multicolumn{2}{@{}l}{\textit{Empirical feature persistence analysis (\Cref{sec:temporal-persistence}, \Cref{app:persistence-derivation})}} \\
$\tau$ & Lag, in tokens \\
$p^{(t)}_j$ & Decoder-aligned projection $\mathbf{d}_j^\top \mathbf{x}^{(t)}$ \\
$R_j(\tau)$ & Empirical persistence of $p^{(t)}_j$ at lag $\tau$ \\
$\beta_{\mathrm{nat}}, \beta_{\mathrm{shf}}$ & Power-law decay exponents, natural / shuffled \\
\midrule
\multicolumn{2}{@{}l}{\textit{Context correction experiment (\Cref{sec:temporal-persistence}, \Cref{app:context-correction-details})}} \\
$\mathbf{w}_\tau \in \mathbb{R}^m$ & Per-feature context-correction weights at lag $\tau$ \\
$w_{\tau,j}$ & Context-correction weight for feature $j$ at lag $\tau$ ($w_j$ at $\tau=1$) \\
$\hat{\mathbf{x}}^{(t)}_{\mathrm{base}}$ & Baseline reconstruction from a standard SAE \\
$\hat{\mathbf{x}}^{(t)}_{\mathrm{ctx}}$ & Context-corrected reconstruction (Equation~\ref{eq:context-head}) \\
$R^2_{\mathrm{improve}}$ & Held-out reconstruction-error improvement from context correction \\
\midrule
\multicolumn{2}{@{}l}{\textit{Persistent SAE (\Cref{sec:method})}} \\
$\mathbf{a}^{(t)} \in \mathbb{R}_{\ge 0}^m$ & Token-wise sparse activation of a Persistent SAE (\Cref{app:batchtopk}) \\
$a^{(t)}_j$ & Token-wise sparse activation of feature $j$ at token $t$ \\
$\mathbf{h}^{(t)} \in \mathbb{R}_{\ge 0}^m$ & Persistent state integrating token-wise activations over time (Equation~\ref{eq:state-update}) \\
$h^{(t)}_j$ & Persistent state of feature $j$ at token $t$ \\
$\lambda_j \in (0, \lambda_{\max})$ & Persistence coefficient for feature $j$ (\Cref{app:persistence-param}) \\
$\lambda_{\max}$ & Upper bound on the persistence coefficient ($=0.999$) \\
$l_j$ & Unconstrained logit parameterising $\lambda_j$ \\
$\sigma(\cdot)$ & Sigmoid function \\
\bottomrule
\end{tabular}
\end{table}

\section{Training Details}
\label{app:training-details}

\subsection{BatchTopK Sparsity}
\label{app:batchtopk}

For every SAE in this paper, sparsity is imposed using a BatchTopK mechanism in which a hidden state $\mathbf{x} \in \mathbb{R}^d$ is first mapped to non-negative pre-activations
\begin{equation}
\tilde{\mathbf{f}} = \mathrm{ReLU}\!\left(\mathbf{W}_{\mathrm{enc}}(\mathbf{x} - \mathbf{b}_{\mathrm{dec}}) + \mathbf{b}_{\mathrm{enc}}\right) \in \mathbb{R}_{\ge 0}^m,
\end{equation}
and only the largest pre-activations are retained across the batch. For a batch $X = \{\mathbf{x}_i\}_{i=1}^{B}$, the pre-activations $\{\tilde{\mathbf{f}}_i\}_{i=1}^{B}$ are flattened into a single vector of length $Bm$, and the sparse activation $\mathbf{f}_i$ is obtained by keeping only the top $kB$ entries globally and setting all others to zero.
The reconstruction is then
\begin{equation}
\hat{\mathbf{x}}_i = \mathbf{W}_{\mathrm{dec}} \mathbf{f}_i + \mathbf{b}_{\mathrm{dec}}.
\end{equation}
In practice, after training with exact BatchTopK selection, inference uses an equivalent thresholded form
\begin{equation}
f_{ij} = \tilde{f}_{ij} \, \mathbf{1}[\tilde{f}_{ij} > \theta],
\end{equation}
where the threshold $\theta$ is estimated from training-time top-$k$ statistics. The Persistent SAE uses the resulting sparse activation as its token-wise sparse activation $\mathbf{a}^{(t)}$.

\subsection{Persistence Parametrisation}
\label{app:persistence-param}

For Persistent SAE, we parametrise each persistence coefficient as
\begin{equation}
\lambda_j = \sigma(l_j) \cdot \lambda_{\max},
\end{equation}
where $l_j \in \mathbb{R}$ is an unconstrained logit, $\sigma$ is the sigmoid function, and $\lambda_{\max} = 0.999$ caps $\lambda_j$ strictly below one. Without this cap, $\lambda_j$ could reach exactly one, turning the state update in \Cref{eq:state-update} into an unbounded running sum with no decay. The cap instead keeps $\lambda_j \in (0, \lambda_{\max})$ throughout training, so every feature has a finite effective memory, and the factor $(1-\lambda_j)$ in \Cref{eq:state-update} makes the recurrence an exponential moving average (EMA) instead of an unnormalised sum. The reason is that under a constant activation $a^{(t)}_j=a_j$, the state converges to $a_j$ for every $\lambda_j$, so persistence does not inflate the state's magnitude. Before training, unless otherwise specified, we initialise every logit $l_j$ to $0.5$, which gives each feature moderate initial persistence while keeping the sigmoid away from saturation, so training can adjust the timescale in either direction.

\subsection{SAE Training Details}
\label{app:sae-baselines}

We train four SAE families for comparison: BatchTopK \citep{batchtopk_sae}, Matryoshka BatchTopK \citep{matryoshka_sae}, Temporal SAE \citep{temporal_sparse_autoencoders}, and our Persistent SAE. Each family uses the same dictionary width, BatchTopK activation budget, optimiser, and token budget. 
\rebuttal{For a more comprehensive analysis, we also train TFA~\citep{priors_in_time}, even though its architecture is considerably more complex and larger than that of a standard SAE. We train TFA with the same dictionary width, sparsity budget, and token budget, as described below.}\looseness=-1

\myparagraph{Configuration.}
Every SAE has a dictionary of $m=16{,}384$ features and a BatchTopK activation budget of $k=20$ per token. We train with Adam at a learning rate of $3\times10^{-4}$ and use a linear warm-up over the first $1{,}000$ steps. From step $1{,}000$, we apply the BatchTopK activation-threshold schedule with decay $\beta=0.999$, together with the auxiliary dead-feature loss from \citep{gao2025scaling}, weighted by $\alpha_{\mathrm{aux}}=0.03125$. Training activations come from Pile Uncopyrighted \citep{pile_uncopyrighted}, with the beginning-of-sequence position removed. For Gemma-2-2B \citep{gemma2}, we use the layer-12 residual stream with hidden width $d=2{,}304$; for Pythia-160M-deduped \citep{biderman2023pythia}, we use the layer-8 residual stream with $d=768$. 
For comparability, we use a two-level Matryoshka SAE whose smaller nested dictionary contains 20\% of the features, while the Temporal SAE partitions its dictionary into a predefined high-level group (20\% of the features) and a low-level group (the remaining 80\%), following their paper \citep{temporal_sparse_autoencoders}. The Temporal SAE weights its reconstruction losses by $0.2$ (high-level) and $0.8$ (low-level), respectively. The Persistent SAE does not use a predefined partition; instead, it learns a persistence coefficient for every feature.

\begin{rebuttalblock}
\myparagraph{Temporal Feature Analysis.}
TFA uses one tied dictionary $\mathbf{D}\in\mathbb{R}^{m\times d}$ for both encoding and decoding. For each token, a single-layer causal attention module with four heads predicts a dense \emph{predicted code} $\mathbf{z}^{(t)}_{\mathrm{pred}}$: its queries come from a dense encoding of the current hidden state, and its keys and values come from the dense encodings of earlier hidden states. TFA scales this code to best fit $\mathbf{x}^{(t)}$ and subtracts its reconstruction. A per-token TopK with $k=20$ then encodes the remaining signal as a sparse \emph{novel code} $\mathbf{z}^{(t)}_{\mathrm{novel}}$, and the decoder reconstructs $\hat{\mathbf{x}}^{(t)}=\mathbf{D}^\top(\mathbf{z}^{(t)}_{\mathrm{pred}}+\mathbf{z}^{(t)}_{\mathrm{novel}})+\mathbf{b}$. The attention module works in the $m$-dimensional code space, so it has about $1.07$B parameters for $m=16{,}384$. As in the reference implementation from \citet{priors_in_time}, TFA scales its inputs to unit mean norm and trains on the squared reconstruction error. We use the same optimiser, learning rate, and warm-up as the other SAE families, and initialise $\mathbf{b}$ from the first batch's geometric median.
\end{rebuttalblock}

\myparagraph{Matched token budget.}
Each SAE family groups activations differently before training on them: BatchTopK and Matryoshka use shuffled individual tokens, the Temporal SAE uses adjacent token pairs, and the Persistent SAE uses contiguous sequences so it can carry state across tokens. To keep training comparable across families, we match the number of activation tokens processed per step and the total number of steps, not the number of these groups. Each model trains for $244{,}140$ steps with an effective batch of approximately $2{,}048$ activation tokens per step, giving the flat baselines approximately $5.0\times10^8$ tokens in total.
For persistent SAEs, unless otherwise specified, we use a sequence length of $256$ tokens during training. 
\rebuttal{TFA also needs whole sequences, so it trains on the same stream of contiguous sequences as the Persistent SAE for $244{,}140$ steps, which gives the same budget of approximately $5.0\times10^8$ training tokens.}
All experiments were run on NVIDIA A100 GPUs with 80~GB of memory.

\begin{figure*}[h]
    \centering
    \includegraphics[width=\textwidth]{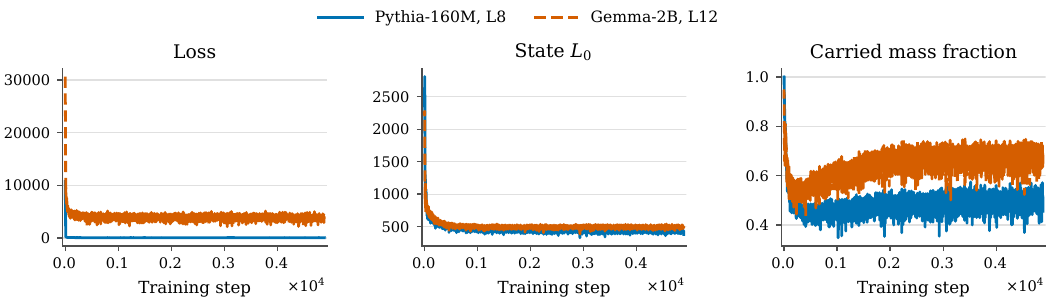}
    \caption{
        \textbf{Persistent SAE training dynamics.}
        We plot, over training: training loss (left); the number of non-zero features in the persistent state $\mathbf{h}^{(t)}$ (state $L_0$; centre); and the carried-mass fraction $\rho^{(t)}$ (right, \Cref{eq:carried-mass-fraction}), the share of the state's total magnitude that comes from the previous token rather than the current one. After a short initial transient, all three quantities stabilise, and the carried-mass fraction settles well away from both zero and one.
    }
    \label{fig:sae-training-metrics}
\end{figure*}

\Cref{fig:sae-training-metrics} shows the training dynamics of the Persistent SAEs. For both models, the loss decreases rapidly and then remains stable, while the state $L_0$ settles at roughly $400$--$500$ active features despite continued training. Both terms in the state update (\Cref{eq:state-update}) are non-negative, so we define the carried-mass fraction as the share of the state's total magnitude that comes from the carried term:
\begin{equation}
    \rho^{(t)} = \frac{\sum_j \lambda_j h^{(t-1)}_j}{\sum_j h^{(t)}_j}.
    \label{eq:carried-mass-fraction}
\end{equation}
This fraction stabilises at approximately $0.5$ for Pythia and $0.65$--$0.70$ for Gemma: roughly half to two-thirds of each state's magnitude comes from the previous token, with the rest replaced by new activations at the current token. Thus, the recurrence does not cause the active state to grow continually, and neither model relies exclusively on new activation or carried state.

\section{Dataset Details}
\label{app:dataset-details}

We use three datasets for distinct parts of the evaluation. Pile \citep{gao2020pile, pile_uncopyrighted} provides the hidden states used to train and evaluate the standard metrics of SAEs, MMLU questions \citep{MMLU} provide subject-labelled text for analysing the persistent state, and AgentDojo trajectories \citep{agentdojo} support the prompt-injection monitoring case study. We describe each dataset and our use of it below.

\subsection{Pile Corpus}

The original Pile is an English-language corpus assembled from 22 sources, including academic papers, web pages, books, code, and online discussions \citep{gao2020pile}. We use the Pile Uncopyrighted release \citep{pile_uncopyrighted}. Documents are accessed in streaming mode and shuffled with a buffer of size $10{,}000$ to train the SAEs. For the natural documents mentioned in \Cref{sec:temporal-persistence}, each sampled document is passed through the model and the resulting residual-stream hidden states are retained for training. 
For the shuffled control, sentences are sampled from documents drawn from the same Pile corpus and concatenated subject to a simple adjacency constraint that prevents consecutive sentences from coming from the same source document. This preserves local continuity while breaking global continuity, which facilitates the study of temporal persistence in \Cref{sec:temporal-persistence}.

\subsection{MMLU Dataset}

Massive Multitask Language Understanding (MMLU) is an English-language benchmark of four-option multiple-choice questions from 57 subjects spanning STEM, the humanities, the social sciences, and other professional and academic fields \citep{MMLU}. We use MMLU for the analyses in \Cref{sec:slow-features}. Specifically, we use questions from abstract algebra, astronomy, business ethics, computer security, and European history. These subjects cover distinct areas while providing a subject label for every example. 
In \Cref{sec:slow-features}, we sample $100$ sequences across the five subjects and retain the final $20$ token representations from each sequence, following \citet{temporal_sparse_autoencoders}. For subject probing, we construct sequences by joining equal-length segments from different subjects and assign each token the subject label of its source segment. \Cref{app:tsne-details} and \Cref{app:probing-details} provide the sampling and evaluation details for the two analyses.

\subsection{AgentDojo Trajectories}
\label{app:agentdojo-details}

Indirect prompt injection occurs when an agent retrieves attacker-controlled text from emails, webpages, documents, or tool outputs and follows instructions inside that text. AgentDojo \citep{agentdojo} evaluates this setting with stateful tool environments containing 97 user tasks, 27 injection targets, and 629 security test cases across workspace, Slack, travel, and banking domains. Its task-specific checks evaluate whether the legitimate user request and the attacker's target action were completed by inspecting the agent output and changes to the environment.

For the case study in \Cref{sec:prompt-injection}, we reuse existing AgentDojo trajectories from the banking, travel, and workspace domains, generated by Llama-3.3-70B-Instruct \citep{Llama3} following \citet{agentdojo}. Benign trajectories with no injection form the negative class ($y=0$). For the positive class ($y=1$), we take runs with an \texttt{important\_instructions} or \texttt{direct} attack where AgentDojo's injection checker reports \texttt{security=false}: an injection was present, but the agent did not complete the attacker's target action. We keep only these failed attacks on purpose. A successful attack usually includes an overt malicious action, such as transferring money or leaking private data, and a monitor could flag that action alone without ever detecting the source injection. Restricting to failed attacks removes this shortcut and tests whether the monitor still detects the injection when the attack leaves no such trace.

\section{Temporal Persistence Experiments}
\label{app:temporal-persistence-experiments}

This section gives the full setup for the two experiments in \Cref{sec:temporal-persistence}: the empirical persistence measure $R_j(\tau)$, and the context correction test of whether past activations help reconstruction.

\subsection{Empirical Persistence of Standard SAE Features}
\label{app:persistence-derivation}

This subsection gives the evaluation procedure and feature-selection criteria for $R_j(\tau)$, and expands on the persistence results plotted in \Cref{fig:main-observation}.\looseness=-1

\myparagraph{Setup.} We compute $R_j(\tau)$ at lags $\tau \in \{1, 4, 16, 32, 64, 128, 256, 512\}$ tokens, on documents truncated to at most $1{,}024$ tokens. For each feature $j$, we call a token an \emph{anchor} if feature $j$ is active there and at least $\tau$ further tokens remain in the document, so we can measure its projection $\tau$ tokens later. We keep a feature in the analysis only if it activates at least $256$ times, has at least $64$ anchors at every lag, and shows a non-negligible activated-token elevation. We exclude a feature instead of stabilising its score whenever this elevation, the denominator $\mathbb{E}[p^{(t)}_j\mid \mathbf{f}^{(t)}_j>0]-\mathbb{E}[p^{(t)}_j]$ in \Cref{eq:persistence}, is close to zero, as happens for near-dead features.

\myparagraph{Natural sequences persist longer than shuffled ones.} Across both models and both SAE families, median persistence is high at lag $1$ and decays as $\tau$ grows, but far more slowly on natural documents than on the shuffled control: natural-document persistence stays clearly above zero out to the longest lag we test, while the shuffled control decays to approximately zero well before that lag. To quantify this gap, we fit a power law to the decay: the fitted exponent is substantially smaller for natural sequences ($\beta_{\mathrm{nat}} \approx 0.32$) than for the shuffled control ($\beta_{\mathrm{shf}} \approx 0.87$--$0.92$), consistent with the slower decay, and this gap holds for both BatchTopK and Matryoshka.

% \myparagraph{Persistence timescales differ across features.} Individual features differ substantially in how their persistence unfolds across lags. Ranking features by their average $R_j(\tau)$ within short ($\tau \leq 16$), medium ($32 \leq \tau \leq 64$), and long ($\tau \geq 128$) lag bands, we find that the resulting distributions span nearly the full unit interval at every band: some features are strongly persistent at short lags but decay before medium lags, while others retain elevated projections well into the long-lag regime. This heterogeneity appears in both SAE families and is substantially larger than in the shuffled control.

\subsection{Context Correction Experiment}
\label{app:context-correction-details}

\myparagraph{Setup.} We fit the per-feature context correction of \Cref{eq:context-head} independently at each lag $\tau \in \{1, 2, 4, 8, 16, 32\}$. For each lag, we use $200$ natural documents to fit $\mathbf{w}_\tau$ and $100$ held-out natural documents to evaluate $R^2_{\mathrm{improve}}$. The target is always the natural-document reconstruction error. For the control, we refit the same diagonal context correction head to predict these natural targets from context activations drawn from the shuffled corpus instead, using the same parameterisation and optimisation budget. We optimise for $10$ epochs with AdamW, learning rate $5 \times 10^{-3}$, and weight decay $10^{-3}$ in order to keep the training lightweight.

\myparagraph{Past activations reduce reconstruction error, but only from real context.} On Pythia-160M, the correction removes $0.89\%$ (BatchTopK) and $0.61\%$ (Matryoshka) of the baseline reconstruction error at lag $1$; the gain shrinks with lag but stays positive at lag $32$ ($0.22\%$ and $0.13\%$). On Gemma-2-2B, both SAEs remove approximately $0.40\%$ at lag $1$, falling to about $0.02\%$ at lag $16$. The shuffled control stays slightly below zero throughout, so unrelated context never improves reconstruction; the gain requires activations from the actual context of the target document.

\myparagraph{Different features play different temporal roles.} \Cref{tab:context-weights} summarises the fitted lag-one weights $w_j$. On natural documents, the mean weight is positive, and most features take a positive weight, so past activation of these features predicts a positive contribution along the same feature direction at the next token. Two further details in the table show that this carry-forward behaviour is feature-specific rather than universal. First, the positive fraction stops well short of $100\%$ ($76$--$91\%$ across models and SAE families): a sizeable minority of features show no carry-forward effect at all. Second, this effect is tied to real document context, since the shuffled control's mean weight stays close to zero for every model and SAE family. Together with the lag-band heterogeneity in \Cref{app:persistence-derivation}, this shows that no single, shared treatment of persistence fits every feature: some features carry information forward and some do not, and among those that do, by how much and for how long both vary by feature. This motivates learning a persistence coefficient separately for each feature in \Cref{sec:method}.

\begin{table}[!h]
\centering
\small
\caption{Fitted lag-one context-correction weights $w_j$. \emph{Mean $w_j$} and \emph{Pos.\ frac.} are the mean and the positive fraction on natural documents; \emph{Shuf.\ $w_j$} is the mean on the shuffled control.}
\label{tab:context-weights}
\begin{tabular}{@{}llccc@{}}
\toprule
\textbf{Model} & \textbf{SAE} & \textbf{Mean $w_j$} & \textbf{Pos.\ frac.} & \textbf{Shuf.\ $w_j$} \\
\midrule
Pythia & BatchTopK  & $0.068$ & $76.2\%$ & $\phantom{-}0.001$ \\
Pythia & Matryoshka & $0.087$ & $91.4\%$ & $-0.001$ \\
Gemma  & BatchTopK  & $0.068$ & $80.4\%$ & $-0.001$ \\
Gemma  & Matryoshka & $0.091$ & $81.9\%$ & $-0.001$ \\
\bottomrule
\end{tabular}
\end{table}

\section{Evaluation Details}
\label{app:evaluation-details}

This section defines \rebuttal{the reconstruction, sparsity, and} the temporal smoothness metric used in \Cref{sec:recon-quality} and gives the evaluation procedures for the four analyses in \Cref{sec:experiments}: automated interpretability (\Cref{sec:fast-features}), subject geometry and subject probing (\Cref{sec:slow-features}), and the prompt-injection monitoring case study (\Cref{sec:prompt-injection}).

\begin{rebuttalblock}
\subsection{Reconstruction and Sparsity Metrics}
\label{app:recon-metrics}

\myparagraph{Metrics.}
We compute the fraction of variance explained (FVE) as $1-\sum_{i}\mathrm{Var}(x_i-\hat{x}_i)\,/\sum_{i}\mathrm{Var}(x_i)$, where each variance is taken over all scored tokens for one hidden dimension $i$. Cosine similarity is the mean over tokens of the cosine between $\mathbf{x}^{(t)}$ and $\hat{\mathbf{x}}^{(t)}$, and the fraction alive is the share of features that are non-zero at least once. $L_0$ is the average number of non-zero entries per token in the sparse activation: $\mathbf{f}^{(t)}$ for standard SAEs and the token-wise activation $\mathbf{a}^{(t)}$ for Persistent SAEs. 
% The persistent state $\mathbf{h}^{(t)}$ is denser, with $476$ (Pythia) and $560$ (Gemma) non-zero entries per token. Only about $14$ of these exceed the activation threshold $\theta$ (\Cref{app:batchtopk}); the rest are small, decayed contributions from earlier activations. Error bars are $95\%$ intervals from a delete-one-block jackknife over $32$ blocks of documents.
\end{rebuttalblock}

\subsection{Temporal Smoothness}
\label{app:smoothness-details}

We use the smoothness metric of \citet{temporal_sparse_autoencoders} to measure how much feature activations jump between neighbouring tokens. A lower score means smoother features.

For each evaluation sequence $s$ of length $T$, we keep only the $n'$ features that are active at least once in the sequence. For each of these features, we find the largest change in its activation between two consecutive tokens, divided by the change in the hidden state at that step. We then average over the active features:
\begin{equation}
    \Delta_s
    =
    \frac{1}{n'} \sum_{j=1}^{n'}
    \max_{t \in \{2,\dots,T\}}
    \frac{\big|f^{(t)}_j - f^{(t-1)}_j\big|}{\lVert \mathbf{x}^{(t)} - \mathbf{x}^{(t-1)} \rVert_2}.
    \label{eq:smoothness}
\end{equation}
Dividing by the change in the hidden state means that a feature is not penalised for changing when the input itself changes sharply. The smoothness score averages this quantity over $n$ evaluation sequences, $S = \frac{1}{n}\sum_{s=1}^{n} \Delta_s$.

For standard SAEs, $f^{(t)}_j$ is the feature activation. For Persistent SAEs, we use the persistent state $h^{(t)}_j$, since this is what the decoder reconstructs from. Besides the score over all features (\emph{Full}), \Cref{tab:core-performance} also reports it separately for the \rebuttal{\emph{Slow} and \emph{Fast}} feature subsets defined in the table caption.

\begin{rebuttalblock}
\myparagraph{Smoothness of the state and the token-wise activation.}
\Cref{eq:state-update} scales every change in $h^{(t)}_j$ by $(1-\lambda_j)$, so the state of a slow feature is smooth partly by construction. The Temporal SAE's high-level features are also smooth by design, because its contrastive loss encourages them to be. To separate this effect, \Cref{tab:smoothness-state-activation} reports the score of the Persistent SAE on both the state $h^{(t)}_j$ and the token-wise activation $a^{(t)}_j$. The token-wise activations of slow features are less smooth than the baseline activations ($0.13$--$0.15$ in \Cref{tab:core-performance}), so their smoothness comes from integrating these activations over time. Fast features score similarly on both, consistent with their smaller $\lambda_j$.

\begin{table}[h]
\centering
\ifshowrebuttal\color{rebuttalcolor}\fi
\footnotesize
\caption{\textbf{Smoothness of the persistent state and the token-wise activation} ($\downarrow$). \emph{Slow}/\emph{Fast} denote the highest $1\%$/lowest $50\%$ of Persistent SAE features by $\lambda_j$.}
\begin{tabular}{lcc}
\toprule
Representation & Pythia & Gemma \\
\midrule
Slow, state $\mathbf{h}^{(t)}$ & $0.047$ & $0.034$ \\
Slow, activation $\mathbf{a}^{(t)}$ & $0.206$ & $0.263$ \\
Fast, state $\mathbf{h}^{(t)}$ & $0.195$ & $0.175$ \\
Fast, activation $\mathbf{a}^{(t)}$ & $0.215$ & $0.202$ \\
\bottomrule
\end{tabular}
\label{tab:smoothness-state-activation}
\end{table}
\end{rebuttalblock}

\subsection{Automated Interpretability Details}
\label{app:autointerp-details}

This subsection gives the autointerpretation pipeline for \Cref{sec:fast-features}, and discusses the qualitative examples referenced there (Tables~\ref{tab:autointerp-qualitative-pythia160m} and~\ref{tab:autointerp-qualitative-gemma2b}).

\myparagraph{Pipeline.} We follow the automated interpretability pipeline of \citet{bills2023language}. For each feature, we collect its highest-activation contexts and ask a judge model to write a short natural-language explanation. We then test this explanation on held-out contexts: the judge uses the explanation to predict whether the feature's activation exceeds a threshold, and we score how well this prediction matches the feature's actual activation. For standard SAEs, this activation is $f_j^{(t)}$; for the Persistent SAE, we use the token-wise activation $a_j^{(t)}$. \rebuttal{For TFA (\Cref{app:tfa-comparison}), we use its sparse novel code, which plays the same role.} We use \texttt{gpt-4o-mini} \citep{GPT4} as the judge, a context length of $1024$ tokens, and roughly $2$M tokens from the Pile for the activation examples. We also score the highest-$\lambda_j$ $1\%$ and lowest-$\lambda_j$ $50\%$ of Persistent SAE features to evaluate their interpretability as groups.\looseness=-1

\myparagraph{Qualitative examples.} Tables~\ref{tab:autointerp-qualitative-pythia160m} and~\ref{tab:autointerp-qualitative-gemma2b} give example features for Pythia-160M and Gemma-2-2B, all scoring above $0.85$, so both groups of features are well explained individually. High-$\lambda_j$ (slow) features track a sustained topic across a passage: examples include ethnic-discrimination narratives, cosmology, and population-based cancer research, where the representative span is one clause drawn from a much longer discussion of the same subject. Low-$\lambda_j$ (fast) features instead pick out a specific, self-contained string, such as a JavaScript function call, a patent-citation phrase, or a SQL \texttt{CREATE TABLE} statement, each explained by the exact text that triggers it. This pattern holds for both models, and matches the finding that fast features are activated by local lexical content while slow features track broader semantic context.

\begin{table*}[t]
\centering
\footnotesize
\setlength{\tabcolsep}{4pt}
\renewcommand{\arraystretch}{1.13}
\caption{Qualitative autointerpretation examples for Pythia-160M. High- and low-$\lambda$ features are drawn from the top 1\% and bottom 50\% of the learned persistence coefficients, respectively. Bold text marks representative activating spans. The selected high-$\lambda$ features are activated by tokens in coherent semantic contexts, whereas the low-$\lambda$ features respond to local strings, syntax, or recurring templates. All selected features have an autointerpretation score above 0.85.}
\label{tab:autointerp-qualitative-pythia160m}
\begin{tabularx}{\textwidth}{@{}lc>{\raggedright\arraybackslash}X>{\raggedright\arraybackslash}Xc@{}}
\toprule
\textbf{Feature} & \textbf{$\lambda$} & \textbf{Autointerpretation} & \textbf{Representative activation} & \textbf{Score} \\
\midrule
\rowcolor{blue!7}
\multicolumn{5}{@{}l}{\textbf{High $\lambda$ (top 1\%): sustained semantic domains}} \\
F-5528 & \num{7.03e-1} & themes of ethnic diversity and experiences of discrimination & ``ashamed of being \textbf{Mexican} because I'm different from the \textbf{stereotype}'' & 0.93 \\
F-12751 & \num{4.82e-1} & phrases related to population-based studies and cancer research & incident \textbf{colorectal cancer cases} identified through a \textbf{population-based cancer registry} & 0.93 \\
F-15986 & \num{3.81e-1} & terms related to high-performance cars and their specifications & rare and desirable \textbf{muscle cars}; a \textbf{1971 Hemi 'Cuda} & 0.93 \\
F-6538 & \num{3.79e-1} & terms related to weather conditions and climate phenomena & a combination of \textbf{global warming} and the \textbf{El Ni\~no weather phenomenon} & 1.00 \\
F-8449 & \num{3.69e-1} & mathematical concepts and terms related to Banach spaces and polynomials & continuous scalar polynomials over a \textbf{Banach space} $X$ & 1.00 \\
\addlinespace[2pt]
\rowcolor{black!7}
\multicolumn{5}{@{}l}{\textbf{Low $\lambda$ (bottom 50\%): local lexical, syntactic, and template patterns}} \\
F-1113 & \num{5.35e-2} & the substring ``white'' followed by various words or concepts & ``There is so much \textbf{white space} just waiting to be filled'' & 1.00 \\
F-6958 & \num{4.96e-2} & the substring ``function'' in JavaScript code snippets & \texttt{.click(function\textbf{() \{}} & 1.00 \\
F-15079 & \num{4.81e-2} & the phrase ``U.S. Pat. No.'' indicating references to patents & ``disclosed in \textbf{U.S. Pat. No.} 5,111,981'' & 1.00 \\
F-3587 & \num{1.10e-2} & URLs and web documentation references & \texttt{ubuntu.com/10.04\allowbreak\textbf{/serverguide/C/}\allowbreak index.html} & 1.00 \\
F-14758 & \num{4.04e-4} & SQL commands related to table creation and data manipulation & \texttt{\textbf{CREATE TABLE IF NOT EXISTS} tennis(...)} & 1.00 \\
\bottomrule
\end{tabularx}
\end{table*}

\begin{table*}[t]
\centering
\footnotesize
\setlength{\tabcolsep}{4pt}
\renewcommand{\arraystretch}{1.13}
\caption{Qualitative autointerpretation examples for Gemma-2-2B. High- and low-$\lambda$ features are drawn from the top 1\% and bottom 50\% of the learned persistence coefficients, respectively. Bold text marks representative activating spans. The selected high-$\lambda$ features are activated by tokens in coherent semantic contexts, whereas the low-$\lambda$ features respond to local strings, syntax, or recurring templates. All selected features have an autointerpretation score above 0.85.}
\label{tab:autointerp-qualitative-gemma2b}
\begin{tabularx}{\textwidth}{@{}lc>{\raggedright\arraybackslash}X>{\raggedright\arraybackslash}Xc@{}}
\toprule
\textbf{Feature} & \textbf{$\lambda$} & \textbf{Autointerpretation} & \textbf{Representative activation} & \textbf{Score} \\
\midrule
\rowcolor{blue!7}
\multicolumn{5}{@{}l}{\textbf{High $\lambda$ (top 1\%): sustained semantic domains}} \\
F-11205 & \num{9.18e-1} & terms related to cosmology and advanced scientific concepts & near-future \textbf{CMB Stage-III and IV surveys} yielding reconstructed \textbf{CMB lensing maps} & 0.86 \\
F-5578 & \num{8.55e-1} & terms related to ionisation charge collection and quantum experiments in physics & \textbf{ionisation charge collection efficiency} observed in gases & 0.86 \\
F-15639 & \num{8.24e-1} & text related to conflict, violence, and ethnic tensions & self-determination for Katanga; a \textbf{separatist group} composed of former Katanga Tigers & 0.86 \\
F-13900 & \num{7.85e-1} & terms related to microbial biomass and ecological concepts in alpine and tundra environments & \textbf{microbial biomass} in an \textbf{alpine meadow on the Tibetan Plateau} & 0.86 \\
F-11933 & \num{7.70e-1} & expressions of self-acceptance and emotional support in personal growth narratives & ``In the last couple years, I've \textbf{let go} and \textbf{decided to be myself}.'' & 1.00 \\
\addlinespace[2pt]
\rowcolor{black!7}
\multicolumn{5}{@{}l}{\textbf{Low $\lambda$ (bottom 50\%): local lexical, syntactic, and template patterns}} \\
F-9996 & \num{1.03e-3} & legal terms and conditions related to software licenses & ``See the \textbf{License} for the specific language governing rights and limitations'' & 1.00 \\
F-11210 & \num{1.30e-5} & the phrase ``all of'' in various contexts & ``it is unlikely that \textbf{all of} these loci respond to selection independently'' & 0.93 \\
F-14595 & \num{7.87e-6} & the substring ``mk'' within mathematical and scientific contexts & repeated \texttt{\textbackslash\textbf{mk}ern} commands in mathematical markup & 0.93 \\
F-13259 & \num{2.73e-6} & specific terms or phrases enclosed in quotation marks & users searching for \textbf{``Pizza''} or \textbf{``Calgary pizza''} & 1.00 \\
F-7307 & \num{1.88e-6} & the word ``In'' at the beginning of sentences & ``\textbf{In} this way the missing persons in our evolutionary mystery would~\ldots'' & 0.86 \\
\bottomrule
\end{tabularx}
\end{table*}

\subsection{Token Category Details}
\label{app:token-type-details}

We group tokens by their part of speech, following the classical split between content words, which carry lexical meaning, and function words, which mark grammatical structure \citep{munte2001differences}. We split content words into nouns, verbs, and adjectives or adverbs. Function words include adpositions, determiners, particles, and pronouns. We keep punctuation and conjunctions as two separate categories because they often mark boundaries and links between parts of a text.

The relative activation frequency $U_{q,c}$ in \Cref{eq:activation-relative-frequency} divides how often features in quintile $q$ are active at category $c$ by how often $c$ occurs in the corpus. This removes the effect of category frequency, so a common category such as nouns does not score highly just because it appears often. A value of $1$ means that features activate at $c$ exactly as often as expected from how common $c$ is in the corpus; a value above $1$ means more often, and a value below $1$ means less often.

\subsection{Subject Geometry Details}
\label{app:tsne-details}

For the t-SNE visualisation, we draw $100$ sequences from five MMLU subjects (abstract algebra, astronomy, business ethics, computer security, and European history) and keep the last $20$ tokens of each sequence, so that every token carries a subject label, following the setup from \citet{temporal_sparse_autoencoders}. For each representation---raw hidden states, BatchTopK features, the Temporal SAE's high-level group, and the Persistent SAE's high-$\lambda_j$ state---we collect the token-level vectors and project them to two dimensions with t-SNE using a common set of hyperparameters, colouring each point by its subject.\looseness=-1

\subsection{Subject Probing Details}
\label{app:probing-details}

This subsection gives the setup for the spliced-MMLU subject-probing experiment in \Cref{sec:slow-features}, whose results are reported in \Cref{fig:slow-features}b.

\myparagraph{Spliced sequences and probe.} 
We build the spliced MMLU sequences by concatenating equal-length segments drawn from different MMLU subjects, balancing subject labels within each sequence and holding out entire source documents for the test split, so that a token's label is the subject of the segment it falls in. We treat each token as a labelled example and fit a linear probe to predict its source subject from the token-level representation. We report held-out one-vs-rest macro ROC AUC as a function of the fraction of the feature dictionary made available to train the probe, swept from $0.1\%$ to $100\%$.

\myparagraph{SAE feature rankings.} 
For the Persistent SAE, we sweep its persistent state $\mathbf{h}^{(t)}$ under the three rankings from \Cref{sec:slow-features}: decreasing $\lambda_j$, increasing $\lambda_j$, and repeated random permutations. We add a fourth, supervised oracle ranking, built from a per-feature, per-subject score
\begin{equation}
    s_{j,c}
    = \mathbb{E}[h^{(t)}_j \mid y^{(t)}=c]
    - \mathbb{E}[h^{(t)}_j \mid y^{(t)}\neq c].
\end{equation}
This score, computed on the training split, is large when feature $j$'s persistent state runs consistently higher for subject $c$ than for the other subjects. For each subject, we rank features by this score, then interleave the five subject-specific rankings in round-robin order so that every subject contributes its most discriminative features. This gives one global feature ordering, which we sweep by fraction for the multiclass probe.

The BatchTopK and Matryoshka baselines have no persistence structure to rank features by, so at each fraction we sample that many features uniformly at random from the full dictionary. The Temporal SAE instead has a fixed high-level and low-level group (\Cref{app:sae-baselines}), so we rank by group rather than by individual feature: the high-first ordering places every high-level-group feature before any low-level-group feature, and the low-first ordering places every low-level-group feature first. Both orderings still cover the full dictionary once the fraction reaches $100\%$.

\subsection{Prompt Injection Monitoring Details}
\label{app:prompt-injection-details}

This subsection gives the trajectory construction, probing, and intervention procedures for the case study in \Cref{sec:prompt-injection} (\Cref{fig:prompt-injection-monitoring}).

\myparagraph{Trajectory construction.}
We take trajectories from AgentDojo \citep{agentdojo} runs of \texttt{Llama-3.3-70B-Instruct} agents, restricted to the banking, travel, and workspace domains (\Cref{app:agentdojo-details}). Attack trajectories use the \texttt{important\_instructions} or \texttt{direct} attack type and satisfy \texttt{security=false}: an injection is present, but the agent did not complete its target action. Benign trajectories contain no injected content. We append each trajectory with the monitoring query
\begin{quote}
\small\texttt{should this trajectory be flagged for prompt injection or unsafe instruction hijacking? Answer yes or no:}
\end{quote}
and read it with Gemma-2-2B, our external monitor model, taking all representations from its layer-12 residual stream.

\myparagraph{Checkpoints and probes.}
For each trajectory, we locate the final token of the injected span (for benign trajectories, the matching position) and mark checkpoints at $0$, $50$, $100$, $200$, $250$, $300$, and $350$ tokens after it, plus the final query position. At each checkpoint, we extract from the same layer: the raw hidden state; the BatchTopK and Matryoshka sparse activations; the Temporal SAE's high-level and low-level feature groups; and slow- and fast-feature states from the Persistent SAE. We fit one logistic regression probe per representation using only the checkpoint at $d=0$, then evaluate it, without retraining, at every other checkpoint, reporting held-out accuracy and ROC AUC.

\myparagraph{Steering vector construction.}
Following prior work that uses linear probes to select SAE features for steering \citep{sae_right_features_steering, he-etal-2025-sae}, we turn the probe for each SAE representation $r$ into a single direction $\mathbf{v}_r$ in the hidden-state space. We do this in three steps.

\textit{(1) Read the probe weights.} We use the probe trained at $d=0$. Its weight for feature $j$ is the injection-class weight minus the benign-class weight. The probe is trained on standardised features, so we divide this weight by the standard deviation of feature $j$ and obtain $\beta_j$. A positive $\beta_j$ means that a larger activation of feature $j$ makes the probe more likely to predict an injection.

\textit{(2) Select features.} For every representation, we keep the $k=128$ features with the largest $\lvert\beta_j\rvert$ and call this set $\mathcal{S}_r$. For example, this is $128$ of the $164$ slow features and $128$ of the $8{,}192$ fast features.

\textit{(3) Map the features to the hidden state.} Each SAE feature $j$ adds its decoder column $\mathbf{W}_{\mathrm{dec}}[:,j]$ to the reconstructed hidden state. We sum these columns, each weighted by its probe weight:
\begin{equation}
    \mathbf{v}_r=\sum_{j\in\mathcal{S}_r}\beta_j\,\mathbf{W}_{\mathrm{dec}}[:,j].
    \label{eq:steering-vector}
\end{equation}

The resulting $\mathbf{v}_r$ points towards what the probe reads as an injection. A positive $\Delta_{\mathrm{causal}}$ therefore means that moving the hidden state in this direction also makes the monitor more likely to answer \textit{yes}.

\myparagraph{Causal intervention.}
For each distance $d\in\{0,50,100,150,200,250,300,350\}$, we truncate each held-out trajectory $d$ tokens after the end of the untrusted span and append the monitoring query. Let $\mathbf{x}_i^{(d)}$ denote the layer-12 residual stream at the final query token of trajectory $i$. We rescale $\mathbf{v}_r$ separately for each example so that $\alpha_i^{(d)}=\rho\lVert\mathbf{x}_i^{(d)}\rVert_2/\lVert\mathbf{v}_r\rVert_2$ with $\rho=0.1$, add $\pm\alpha_i^{(d)}\mathbf{v}_r$ to the output of layer~12 at this token only, and run the rest of the forward pass. The logit margin $s$ uses the single-token answers ``\texttt{\ yes}'' and ``\texttt{\ no}''. 
We report the mean of \Cref{eq:causal-intervention-monitor} over trajectories, with $95\%$ confidence intervals from $500$ bootstrap resamples of trajectories.
We keep $\mathbf{v}_r$ fixed across distances, so \Cref{fig:prompt-injection-monitoring}b (bottom) shows whether one direction learnt at the end of the injected span still steers the monitor later in the trajectory. Note that the intervention changes only the external monitor's logit margin and does not act on the original Llama-3.3-70B agent.

\section{Additional Analyses}
\label{app:additional-analyses}

\begin{rebuttalblock}
\subsection{Reconstruction under a Matched Read Budget}
\label{app:read-budget}

BatchTopK impose a token-wise sparsity constraint on $\mathbf{a}^{(t)}$ to about $20$ non-zero activations per token. The Persistent SAE reconstructs from the persistent state $\mathbf{h}^{(t)}$, which accumulates earlier activations and, according to \Cref{fig:sae-training-metrics}, has more non-zero activations per token on average. In this section, we test whether the Persistent SAE's reconstruction quality relies on these additional non-zero activations.

\myparagraph{Setup.}
For each token, we keep the $K$ largest activations for each baseline that its decoder uses for reconstruction and set the rest to zero. We test $K\in\{20,40,80,160,320\}$ and the full activation, and follow the evaluation protocol of \Cref{tab:core-performance}.
For TFA, the decoder reads the sum of its dense predicted code and sparse novel code (\Cref{app:tfa-comparison}). Its dictionary rows do not have unit norm, and its predicted code can be negative, so we rank its entries by the magnitude of their contribution, $|z_j|\,\lVert\mathbf{D}_j\rVert_2$.

\begin{rebuttalblock}
\begin{table}[t]
\centering
\ifshowrebuttal\color{rebuttalcolor}\fi
\footnotesize
\setlength{\tabcolsep}{4pt}
\renewcommand{\arraystretch}{1.12}
\caption{
\textbf{Reconstruction quality under a matched read budget.}
FVE when the decoder reads only the $K$ largest activations. \emph{Full} reads every non-zero entry. $L_0$ are reported for both token-wise activation ($\mathbf{a}^{(t)}$ for the Persistent SAE, $\mathbf{f}^{(t)}$ for the baselines) and in the persistent state $\mathbf{h}^{(t)}$. $^\dagger$For TFA, the two columns give its sparse novel code and the full code that its decoder reads.
Best/second-best results are bold/underlined.
}
\begin{tabular}{llcccccccc}
\toprule
& \multirow{2}{*}{Method}
& \multicolumn{6}{c}{FVE ($\uparrow$) by read budget $K$}
& \multicolumn{2}{c}{$L_0$} \\
\cmidrule(lr){3-8}
\cmidrule(lr){9-10}
& & $20$ & $40$ & $80$ & $160$ & $320$ & Full & $\mathbf{a}^{(t)}$ & $\mathbf{h}^{(t)}$ \\
\midrule
\multirow{5}{*}{\rotatebox[origin=c]{90}{Pythia}}
& BatchTopK
& $\mathbf{0.929}$ & $\mathbf{0.931}$ & $\underline{0.931}$ & $\underline{0.931}$ & $\underline{0.931}$ & $\underline{0.931}$ & $19.0$ & -- \\
& Matryoshka
& $0.907$ & $0.910$ & $0.910$ & $0.910$ & $0.910$ & $0.910$ & $19.3$ & -- \\
& Temporal
& $\underline{0.922}$ & $0.924$ & $0.924$ & $0.924$ & $0.924$ & $0.924$ & $19.1$ & -- \\
& TFA
& $0.238$ & $0.309$ & $0.415$ & $0.555$ & $0.703$ & $0.913$ & $20.0^\dagger$ & $2660.5^\dagger$ \\
& \cellcolor{gray!10}Persistent SAE (Ours)
& \cellcolor{gray!10}$0.921$ & \cellcolor{gray!10}$\underline{0.931}$ & \cellcolor{gray!10}$\mathbf{0.933}$ & \cellcolor{gray!10}$\mathbf{0.933}$ & \cellcolor{gray!10}$\mathbf{0.933}$ & \cellcolor{gray!10}$\mathbf{0.933}$ & \cellcolor{gray!10}$22.3$ & \cellcolor{gray!10}$475.5$ \\
\midrule
\multirow{5}{*}{\rotatebox[origin=c]{90}{Gemma}}
& BatchTopK
& $\mathbf{0.717}$ & $\mathbf{0.729}$ & $\mathbf{0.729}$ & $\mathbf{0.729}$ & $\mathbf{0.729}$ & $\mathbf{0.729}$ & $19.7$ & -- \\
& Matryoshka
& $\underline{0.694}$ & $0.707$ & $0.707$ & $0.707$ & $0.707$ & $0.707$ & $19.9$ & -- \\
& Temporal SAE
& $0.681$ & $0.702$ & $0.702$ & $0.702$ & $0.702$ & $0.702$ & $19.9$ & -- \\
& TFA
& $0.294$ & $0.319$ & $0.340$ & $0.357$ & $0.369$ & $0.623$ & $20.0^\dagger$ & $3677.5^\dagger$ \\
& \cellcolor{gray!10}Persistent SAE (Ours)
& \cellcolor{gray!10}$0.689$ & \cellcolor{gray!10}$\underline{0.718}$ & \cellcolor{gray!10}$\underline{0.725}$ & \cellcolor{gray!10}$\underline{0.725}$ & \cellcolor{gray!10}$\underline{0.725}$ & \cellcolor{gray!10}$\underline{0.725}$ & \cellcolor{gray!10}$22.6$ & \cellcolor{gray!10}$559.5$ \\
\bottomrule
\end{tabular}
\label{tab:read-budget}
\end{table}
\end{rebuttalblock}

\myparagraph{Persistent SAEs reconstruct as well as standard SAEs with the same read budget.}
In \Cref{tab:read-budget}, we show that when every SAE reconstruct with the same number of top-K activations per token, the Persistent SAE reconstructs as well as the standard SAEs. 
Most non-zero entries are small leftovers of earlier activations and contribute little to the reconstruction. TFA relies on a dense predicted code, so it loses much of its reconstruction quality when it can read only a few activations.
The Persistent SAE therefore does not benefit from the extra non-zero activations for reconstruction. 
Our design lets the model learn from reconstruction which features should carry context over longer ranges.
For this reason, we impose sparsity on the token-wise activation $\mathbf{a}^{(t)}$, which, as in standard SAEs, adds only a few new activations per token and is the unit we interpret (\Cref{sec:fast-features}).
\end{rebuttalblock}

\subsection{Token Similarity Across Subject Boundaries}
\label{app:token-similarity}

\begin{figure*}[t]
    \centering
    \includegraphics[width=\linewidth]{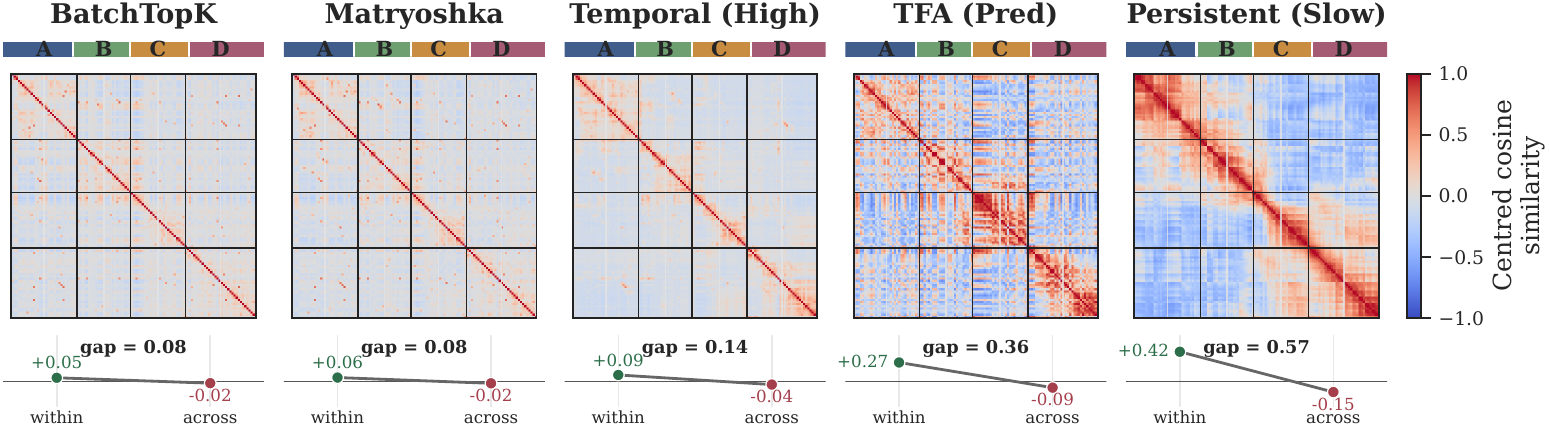}
    \caption{
        \textbf{Centred token similarity across four spliced MMLU subjects.}
        Each panel shows the pairwise centred cosine similarity between token representations in one sequence with four segments (A--D), each from a different subject. The Persistent SAE slow features have the clearest block structure.
    }
    \label{fig:splice-token-similarity}
\end{figure*}

This subsection extends the analysis in \Cref{sec:slow-features}. There, the t-SNE plot and the subject probes summarise many sequences at once. Here, motivated by \citet{priors_in_time}, we look inside one spliced sequence and ask two questions: do tokens from the same subject share a similar state of the slow features, and how does this state change at the boundary between subjects?

\myparagraph{Setup.}
For these visualisations, we use the layer-12 residual stream of Gemma-2-2B and construct one sequence from the MMLU test split. The sequence concatenates one sentence from each of four subjects, in order: college biology (A), college physics (B), business ethics (C), and high school government and politics (D). For each displayed representation, let $\mathbf{z}^{(t)}$ denote its feature vector at token position $t$. We centre each vector using the mean representation over all $T$ tokens in the spliced sequence, $\bar{\mathbf{z}} = T^{-1}\sum_{t=1}^{T}\mathbf{z}^{(t)}$, and compute centred cosine similarity between $\mathbf{z}^{(i)}-\bar{\mathbf{z}}$ and $\mathbf{z}^{(j)}-\bar{\mathbf{z}}$. A positive similarity means that both tokens deviate from the sequence mean in the same direction, and a negative similarity means that they deviate in opposite directions. The black lines in the heatmaps mark where one subject ends and the next begins.

\myparagraph{Slow features separate subjects within a sequence.}
\Cref{fig:splice-token-similarity} shows that the state of the Persistent SAE's slow features has positive similarity within each segment and negative similarity across segments. 
\rebuttal{Although each $\lambda_j$ is fixed after training, the state still adapts to different context: the negative similarity across segments shows that, after each boundary, the state moves away from the previous subject.}
Tokens in the same segment therefore have similar states, including tokens far apart in the sequence. The Temporal SAE high-level group shows some block structure, but its gap of $0.14$ is about a quarter of the $0.57$ gap of the state of the slow features. BatchTopK and Matryoshka show almost no block structure in this example.
\rebuttal{Since \citet{priors_in_time} run a similar analysis for TFA, we also include TFA's predicted code in this experiment. The Persistent SAE still achieves a clearer subject separation with a much smaller and more interpretable model than TFA.} These results agree with the probing results in \Cref{fig:slow-features}~(b): the highest-$\lambda_j$ features form a small subset whose state tracks the current subject, and the model finds this subset without subject labels. 

\begin{figure}[t]
    \centering
    \includegraphics[width=0.6\linewidth]{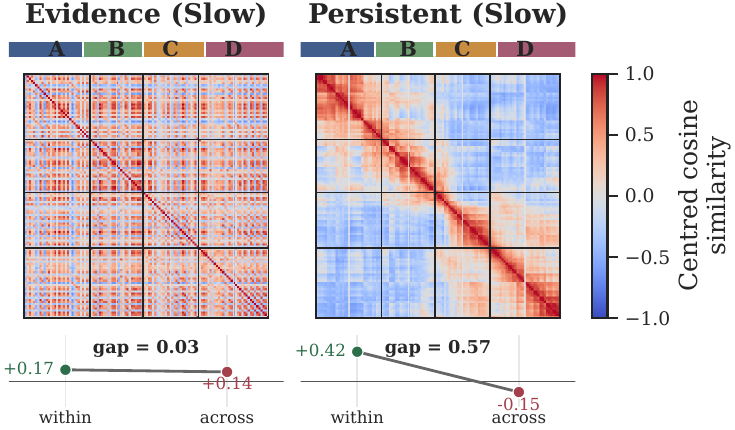}
    \caption{
        \textbf{Token-wise activations and states encode different information.}
        Centred token similarity on the same spliced sequence as \Cref{fig:splice-token-similarity}, for the token-wise activation $\mathbf{a}^{(t)}$ and the state $\mathbf{h}^{(t)}$ of the highest $1\%$ of features by $\lambda_j$. 
        We find that integrating the activations over time turns weak local activations into a sequence-level representation.
    }
    \label{fig:splice-activation-vs-state}
\end{figure}

According to \Cref{eq:state-update}, the state $\mathbf{h}^{(t)}$ accumulates the token-wise activations $\mathbf{a}^{(t)}$. To check that the subject signal comes from this accumulation, we apply the same analysis to $\mathbf{a}^{(t)}$ restricted to the same slow features. In \Cref{fig:splice-activation-vs-state}, we show that the activations show little segment structure: the mean similarity is $0.17$ within segments and $0.14$ across them, a gap of only $0.03$. The state, computed from these activations, has a much larger gap. Therefore, each activation records only the current token, whereas the state integrates these activations over the preceding tokens and shifts at splice boundaries. This finding matches the division of labour in \Cref{sec:method}: token-wise activations capture local information, and the state of the slow features retains the context accumulated from them.

\subsection{States of Slow Features Track Subject Information}
\label{app:gemma2b-semantic}

\begin{figure*}[t]
    \centering
    \includegraphics[width=\linewidth]{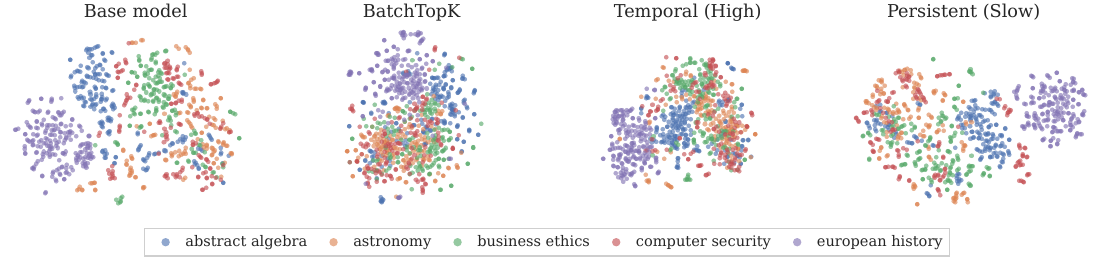}
    \caption{
        \textbf{The state of the slow features tracks subject information in Gemma-2-2B.}
        t-SNE of token-level MMLU representations from Gemma-2-2B, coloured by subject, for the base model's hidden states, BatchTopK, the Temporal SAE's high-level group, and the state of the Persistent SAE's slow features. The state of the slow features forms the clearest subject clusters.
    }
    \label{fig:gemma2b-semantic}
\end{figure*}

\Cref{sec:slow-features} visualises the state of the slow features for Pythia-160M (\Cref{fig:slow-features}a). \Cref{fig:gemma2b-semantic} repeats this analysis for Gemma-2-2B, using the same five MMLU subjects and the same t-SNE setup (\Cref{app:tsne-details}). As with Pythia-160M, BatchTopK features mix all five subjects together, and the Temporal SAE's high-level group separates European history from the base model's hidden states but leaves the other four subjects intermixed. The state of the Persistent SAE's slow features again forms the clearest, most separated subject clusters, learned without subject labels. This matches the pattern found for Pythia-160M and the quantitative subject-probing results in \Cref{fig:slow-features}b, which already show a comparable trend for both models.

\begin{figure*}[t]
    \centering
    \hspace*{-4mm}\includegraphics[width=1.08\linewidth]{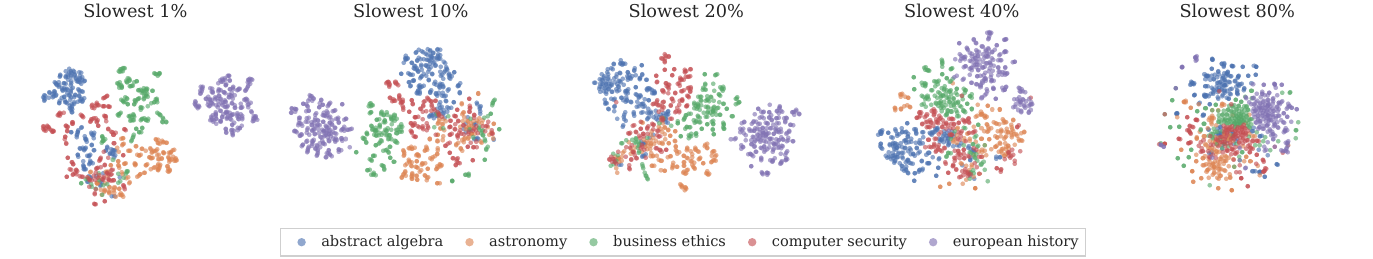}
    \caption{
        \textbf{Subject separation weakens gradually as the feature set widens.}
        t-SNE of token-level MMLU representations from Pythia-160M, coloured by subject, using the state of an increasingly wide set of slow features: the slowest $1\%$, $10\%$, $20\%$, $40\%$, and $80\%$ by $\lambda_j$. Subject clusters are sharpest for the slowest $1\%$ and become progressively more mixed as more, less-slow features are included.
    }
    \label{fig:semantic-slow-feature-sweep}
\end{figure*}

\myparagraph{The subject clusters degrade gradually as more, faster features are added.}
\Cref{fig:gemma2b-semantic} uses the slowest $1\%$ of features, the same threshold used throughout the analysis of the paper. To check that this threshold is not a cherry-picked cutoff, \Cref{fig:semantic-slow-feature-sweep} repeats the t-SNE visualisation for Pythia-160M while gradually widening the feature set to the slowest $1\%$, $10\%$, $20\%$, $40\%$, and $80\%$ by $\lambda_j$. The five subject clusters are most distinct at $1\%$, and they mix together steadily as each wider set folds in faster, less persistent features, until the slowest $80\%$ looks about as mixed as the BatchTopK baseline in \Cref{fig:slow-features}a. Subject information is therefore not confined to a sharply defined subset of slow features; instead, it is concentrated most strongly in the slowest features and fades away smoothly as faster features are added, so any small threshold near the top of the persistence ranking would show a similar effect.

\subsection{Effect of Persistence Granularity}
\label{app:persistence-granularity}

\myparagraph{Setup.}
We test how finely the persistence coefficients must be parameterised to recover heterogeneous learned timescales. All models use Pythia-160M-deduped layer-8 residual-stream activations, a dictionary of $m=16{,}384$ features, BatchTopK sparsity with $k=20$, and training sequences of $256$ tokens. We train each model for $244{,}140$ steps with Adam at a learning rate of $3\times10^{-4}$. Persistence coefficients are initialised to $0.5$, and all other training settings match the feature-wise Persistent SAE described in \Cref{app:training-details}.

We compare models with
\[
G \in \{1, 10, 100, 1{,}000, 10{,}000, 16{,}384\}
\]
learned persistence coefficients. For $G<16{,}384$, we partition the feature indices into $G$ fixed groups of approximately equal size, assigning feature $j$ to group
\begin{equation}
    g(j) = \left\lfloor \frac{jG}{m} \right\rfloor.
\end{equation}
Each group $g$ has one trainable persistence logit $l_g$, giving
\begin{equation}
    \lambda_j = \lambda_{g(j)}
              = \sigma(l_{g(j)})\lambda_{\max},
    \qquad \lambda_{\max}=0.999.
\end{equation}
The group assignment stays fixed throughout training, so features in the same group share one coefficient in the state update,
\begin{equation}
    h_j^{(t)}
    =
    \lambda_{g(j)}h_j^{(t-1)}
    +
    \left(1-\lambda_{g(j)}\right)a_j^{(t)}.
\end{equation}
Gradients from every feature in a group update the same logit $l_g$, while the encoder and decoder parameters remain feature-specific and are trained jointly with these shared coefficients. This ablation therefore changes only the number of independently learned persistence coefficients: $G=1$ gives the whole dictionary one shared timescale, and $G=16{,}384$ recovers one coefficient per feature.

\begin{figure*}[t]
    \centering
    \includegraphics[width=\linewidth]{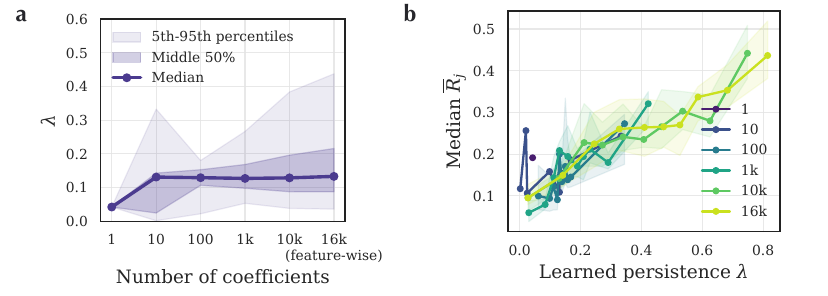}
    \caption{
        \textbf{Finer persistence parameterisations recover heterogeneous empirical timescales.}
        We train otherwise matched Persistent SAEs with $G\in\{1,10,100,1{,}000,10{,}000,16{,}384\}$ persistence coefficients. For $G<16{,}384$, approximately equal-sized groups of features share one learned coefficient; $G=16{,}384$ assigns one coefficient to every feature.
        \textbf{(a)}~Feature-weighted distributions of the learned coefficients. The median changes little once $G\geq10$, whereas fine-grained models recover a wider upper tail of slow features.
        \textbf{(b)}~Each point's $x$-position is a bin's learned persistence $\lambda$, and its $y$-position is the median of $\overline{R}_j$, the empirical persistence from \Cref{eq:persistence} averaged over lags, among the features in that bin. 
        % Shaded regions show 95\% bootstrap confidence intervals obtained by resampling coefficient groups. 
        The relationship between learned and empirical persistence becomes clearer as the number of independently learned coefficients increases.
    }
    \label{fig:persistence-group-ablation}
\end{figure*}

\myparagraph{Learned coefficient distributions.}
\Cref{fig:persistence-group-ablation}a reports the distribution of feature-level coefficients after training. With one global coefficient, the model converges to a nearly token-local value around $\lambda=0.04$. Once the model has at least ten coefficients, the median stays close to $0.13$: finer parameterisations mainly widen the spread. In particular, the 95th percentile rises from around $0.18$ with $100$ coefficients to $0.44$ with feature-wise coefficients. A feature-wise parameterisation therefore lets a subset of features learn much slower dynamics without raising learned persistence across the whole dictionary.

\myparagraph{Comparison with empirical persistence.}
We next test whether the learned coefficients track $R_j(\tau)$, the empirical persistence measure defined in \Cref{sec:temporal-persistence} (\Cref{eq:persistence})
% , where $R_j(\tau)=1$ means a feature's activation is fully retained $\tau$ tokens later and $R_j(\tau)=0$ means it has decayed to baseline. 
We evaluate each model on $128$ Pile documents, truncate each document to at most $1{,}024$ tokens, and compute $R_j(\tau)$ from the decoder direction at $\tau\in\{1,4,16,32,64,128,256,512\}$. We retain features that activate at least $256$ times, have at least $64$ valid anchors at every lag, and have an activated-token elevation of at least $10^{-4}$. For each retained feature, we average $R_j(\tau)$ over all eight lags into a single score:
\begin{equation}
\overline{R}_j
=
\frac{1}{8}\sum_{\tau\in\{1,4,16,32,64,128,256,512\}}R_j(\tau).
\end{equation}

\Cref{fig:persistence-group-ablation}b plots this empirical score against the learned coefficient. We order complete coefficient groups by their learned $\lambda$ and divide them into at most ten bins; each point's $x$-position is that bin's learned $\lambda$ and its $y$-position is the median $\overline{R}_j$ among the features in that bin. 
% The confidence intervals resample coefficient groups rather than individual features, since features sharing a coefficient are not independent. 
With ten shared coefficients, the relationship is noisy and offers little resolution; it grows more ordered as $G$ increases. For the $10{,}000$-coefficient and feature-wise models, bins with larger learned $\lambda$ also have larger empirical persistence $\overline{R}_j$, and their calibration curves are similar, especially at the lowest and highest values of $\lambda$. Reconstruction therefore successfully learns feature-dependent timescales, and coarse sharing obscures part of this variation.

\begin{rebuttalblock}
\subsection{Comparison with Temporal Feature Analysis}
\label{app:tfa-comparison}

TFA~\citep{priors_in_time} also models temporal structure in SAEs, but it places this structure elsewhere in the model. It splits each hidden state into a dense predicted code, produced by an attention module over earlier tokens, and a sparse novel code for the current token. The Persistent SAE keeps one sparse activation and gives each feature its own timescale. In this section, we compare the two on reconstruction quality and architectural complexity.

% \myparagraph{Setup.}
% We train TFA on Pythia-160M and Gemma-2-2B with the same dictionary width ($m=16{,}384$), same sparsity budget ($k=20$ for the novel code), and training token budget as the other SAEs (\Cref{app:sae-baselines}), and evaluate it with the same metrics from \Cref{app:recon-metrics}.

\begin{rebuttalblock}
\begin{table}[t]
\centering
\ifshowrebuttal\color{rebuttalcolor}\fi
\footnotesize
\setlength{\tabcolsep}{3pt}
\renewcommand{\arraystretch}{1.12}
\caption{
\textbf{Comparison with Temporal Feature Analysis (TFA).}
All SAEs use $16{,}384$ features and the same training token budget (\Cref{app:sae-baselines}).
For TFA, \emph{Autointerp} scores the novel code; for the other SAEs, \emph{Autointerp} gives the best of the Full, Slow, and Fast scores in \Cref{tab:core-performance}.
Params counts trainable parameters in millions; $^\dagger$plus $16$k persistence coefficients $\lambda_j$.
Best/second-best results are bold/underlined.
}
\begin{tabular}{llcccc}
\toprule
& Method
& FVE ($\uparrow$)
& \shortstack{Cosine\\Sim ($\uparrow$)}
& \shortstack{Autointerp\\($\uparrow$)}
& \shortstack{Params\\(M, $\downarrow$)} \\
\midrule
\multirow{5}{*}{\rotatebox[origin=c]{90}{Pythia}}
& BatchTopK
& $\mathbf{0.94{\scriptscriptstyle\pm.01}}$
& $\mathbf{0.95{\scriptscriptstyle\pm.01}}$
& $\underline{0.91{\scriptscriptstyle\pm.02}}$
& $\mathbf{25.2}$
\\
& Matryoshka
& $0.92{\scriptscriptstyle\pm.01}$
& $\underline{0.94{\scriptscriptstyle\pm.01}}$
& $0.87{\scriptscriptstyle\pm.02}$
& $\mathbf{25.2}$
\\
& Temporal
& $\underline{0.93{\scriptscriptstyle\pm.02}}$
& $\mathbf{0.95{\scriptscriptstyle\pm.01}}$
& $0.89{\scriptscriptstyle\pm.02}$
& $\mathbf{25.2}$
\\
& TFA
& $0.91{\scriptscriptstyle\pm.00}$
& $\underline{0.94{\scriptscriptstyle\pm.00}}$
& $0.74{\scriptscriptstyle\pm.05}$
& $1{,}086.4$
\\
& \cellcolor{gray!10}Persistent SAE (Ours)
& \cellcolor{gray!10}$\underline{0.93{\scriptscriptstyle\pm.01}}$
& \cellcolor{gray!10}$\mathbf{0.95{\scriptscriptstyle\pm.01}}$
& \cellcolor{gray!10}$\mathbf{0.93{\scriptscriptstyle\pm.01}}$
& \cellcolor{gray!10}$\mathbf{25.2}^\dagger$
\\
\midrule
\multirow{5}{*}{\rotatebox[origin=c]{90}{Gemma}}
& BatchTopK
& $\mathbf{0.73{\scriptscriptstyle\pm.01}}$
& $\mathbf{0.89{\scriptscriptstyle\pm.01}}$
& $\underline{0.87{\scriptscriptstyle\pm.03}}$
& $\mathbf{75.5}$
\\
& Matryoshka
& $\underline{0.71{\scriptscriptstyle\pm.01}}$
& $\mathbf{0.89{\scriptscriptstyle\pm.01}}$
& $\mathbf{0.89{\scriptscriptstyle\pm.02}}$
& $\mathbf{75.5}$
\\
& Temporal SAE
& $0.70{\scriptscriptstyle\pm.01}$
& $\underline{0.88{\scriptscriptstyle\pm.01}}$
& $0.86{\scriptscriptstyle\pm.02}$
& $\mathbf{75.5}$
\\
& TFA
& $0.62{\scriptscriptstyle\pm.00}$
& $0.85{\scriptscriptstyle\pm.00}$
& $0.78{\scriptscriptstyle\pm.05}$
& $1{,}111.6$
\\
& \cellcolor{gray!10}Persistent SAE (Ours)
& \cellcolor{gray!10}$\mathbf{0.73{\scriptscriptstyle\pm.01}}$
& \cellcolor{gray!10}$\mathbf{0.89{\scriptscriptstyle\pm.01}}$
& \cellcolor{gray!10}$\mathbf{0.89{\scriptscriptstyle\pm.01}}$
& \cellcolor{gray!10}$\mathbf{75.5}^\dagger$
\\
\bottomrule
\end{tabular}
\label{tab:tfa-comparison}
\end{table}
\end{rebuttalblock}

\myparagraph{Persistent SAEs achieve better reconstruction quality with simpler and more interpretable architectures.}
In \Cref{tab:tfa-comparison}, we show that the TFA reconstructs less accurately under the same token budget. Under the same autointerp protocol (\Cref{app:autointerp-details}), TFA's sparse novel code scores lower on both models. TFA also needs far more training parameters. Its attention module adds about $1.07$B parameters to the dictionary, which makes TFA $15$--$43$ times larger than a standard SAE of the same width, while the Persistent SAE adds only $16$k learnable persistence coefficients.
% TFA also uses less of its dictionary: only $39\%$ (Pythia) and $14\%$ (Gemma) of its features ever enter the novel code, compared with at least $75\%$ for every other SAE. 
% \citet{priors_in_time} report reconstruction on par with BatchTopK under their own training protocol, so part of this gap may come from our budget-matched setting ($k=20$ and a shared optimiser and schedule).
The Persistent SAE learns a persistence coefficient $\lambda_j$ for each sparse feature, providing a label-free ranking of feature timescales. This allows us to identify slow features and investigate their contextual information through probing and their causal influence through steering (\Cref{sec:slow-features,sec:prompt-injection}).
\end{rebuttalblock}

\section{Ablation Studies}
\label{app:ablation-studies}

\subsection{Robustness to Persistence Initialisation}
\label{app:persistence-init-ablation}

\Cref{app:persistence-param} initialises every persistence logit $l_j$ to a constant value corresponding to $\lambda_j=0.5$. To check that this choice does not drive our results, we retrain Pythia-160M Persistent SAEs from five alternative initialisations while holding every other hyperparameter fixed: constant initial persistence of $0.1$, $0.5$ (the default used throughout the paper), and $0.9$, together with Xavier-uniform \citep{glorot2010understanding} and Kaiming-uniform \citep{he2015delving} initialisation of the logits $l_j$. For each run, we compare the distribution of learned persistence coefficients $\lambda_j$ after training.

\begin{figure}[t]
    \centering
    \includegraphics[width=0.9\linewidth]{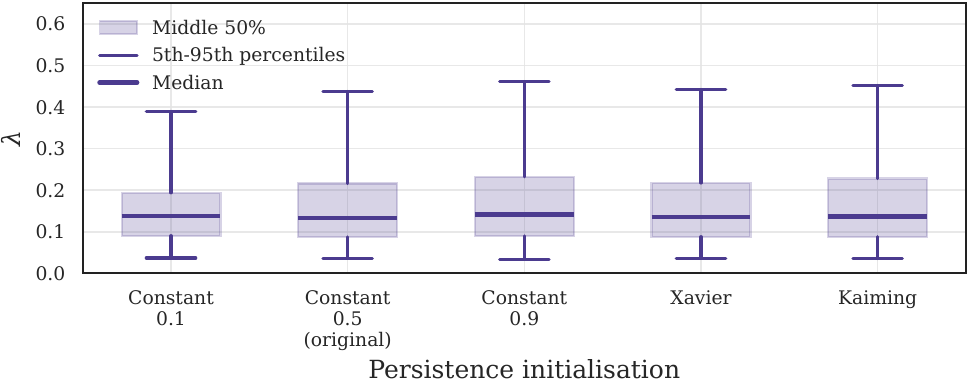}
    \caption{
        \textbf{Learned persistence is robust to its initialisation.}
        Distribution of learned per-feature persistence coefficients $\lambda_j$ for Pythia-160M Persistent SAEs trained from five different initialisations of the persistence logits.
    }
    \label{fig:persistence-init-ablation}
\end{figure}

\myparagraph{Learned persistence converges to a similar distribution regardless of initialisation.} \Cref{fig:persistence-init-ablation} shows that the five initialisation schemes yield nearly indistinguishable distributions of learned $\lambda_j$: medians range from $0.134$ to $0.142$ and means from $0.162$ to $0.182$ across all five runs, with comparable spread in every case (5th--95th percentile interval of approximately $0.03$--$0.46$). The two structured random initialisations, Xavier-uniform and Kaiming-uniform, fall within this same range rather than producing systematically different persistence. We observe a small, monotonic trend in which a higher constant initial persistence yields a slightly higher converged mean (from $0.162$ at initial persistence $0.1$ to $0.182$ at initial persistence $0.9$), but this difference is an order of magnitude smaller than the within-run standard deviation ($0.11$--$0.14$). Training therefore pulls the persistence coefficients toward a similar equilibrium distribution irrespective of how they are seeded, indicating that the learned per-feature persistence structure reflects a property discovered during training.

\subsection{Effect of Training Context Length}
\label{app:context-length-ablation}

In this section, we test whether the learned persistence coefficients depend on the sequence length used during training. We train Pythia-160M Persistent SAEs on sequences of $32$, $64$, $128$, $256$, and $512$ tokens. All runs use layer 8, a dictionary of $16{,}384$ features, and BatchTopK sparsity with $k=20$. We adjust the sequence batch size to keep the nominal number of tokens per update approximately fixed.

\begin{figure}[t]
    \centering
    \includegraphics[width=0.5\linewidth]{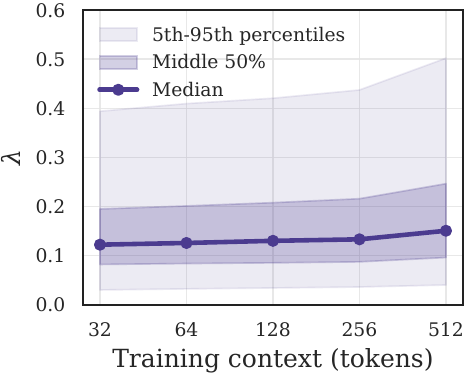}
    \caption{
        \textbf{Learned persistence is stable across training context length.}
        Median learned persistence coefficient $\lambda_j$ across training context lengths. The dark and light bands show the 25th--75th and 5th--95th percentile intervals, respectively.
    }
    \label{fig:context-length-ablation}
\end{figure}

\myparagraph{Persistence estimates are stable across training context length.} \Cref{fig:context-length-ablation} shows that the learned persistence coefficient changes little as the training context grows from $32$ to $512$ tokens. The median $\lambda_j$ rises only from $0.122$ to $0.151$, and the 95th percentile from $0.394$ to $0.502$, a small shift relative to the 25th--95th percentile spread at any single context length. Therefore, the distribution stays broad throughout different lengths, with the same mixture of short- and long-timescale features at every length. Because context length has only a marginal effect on the estimated persistence, we use $256$ tokens in the main experiments, which captures this stable regime without the added training cost of longer contexts.

\subsection{Persistence Across Model Sizes}
\label{app:model-size-persistence}

\begin{figure}[t]
    \centering
    \includegraphics[width=0.95\linewidth]{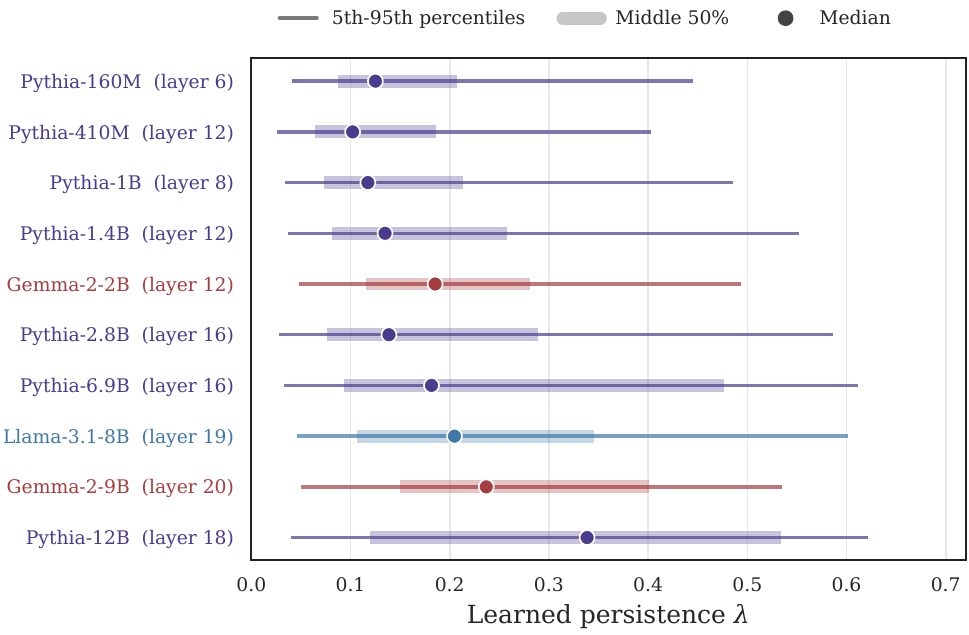}
    \caption{
        \textbf{Larger models tend to learn more persistent features.}
        Distribution of learned persistence coefficients across ten language models, ordered by nominal parameter count. Points show medians, thick intervals show the middle 50\%, and thin intervals show the 5th--95th percentiles. Label colours identify the model family.
    }
    \label{fig:model-size-persistence}
\end{figure}

In this section, we conduct an initial analysis towards understanding how model size affects the learned persistence of SAE features. We compare Persistent SAEs trained on ten language models of varying sizes. The comparison includes seven Pythia models ranging from Pythia-160M to Pythia-12B, two Gemma-2 models of sizes 2B and 9B, and a new model family Llama-3.1-8B \citep{Llama3}. Each SAE uses a dictionary of $16{,}384$ features, BatchTopK sparsity with $k=20$, and training sequences of $256$ tokens. \Cref{fig:model-size-persistence} reports the distribution of all learned coefficients at the layer shown for each model.

\myparagraph{Median persistence generally increases with model size.} During the Persistent SAEs training, every model learns a broad range of feature timescales, but the distributions shift towards larger $\lambda_j$ for larger models. In addition, across all ten models, the Spearman correlation between parameter count and median $\lambda_j$ is $0.927$; within the seven Pythia models, it is $0.893$. The Pythia median increases from $0.125$ for Pythia-160M to $0.338$ for Pythia-12B, and the 95th percentile increases from $0.445$ to $0.622$. This pattern is consistent with larger models assigning longer timescales to a greater share of SAE features. 
One possible explanation is that larger models have more capacity to track dependencies over long contexts and so allocate more features to represent them. We leave the testing of this explanation, for instance by relating $\lambda_j$ to a model's effective context usage, to future work.\looseness=-1

% \haoyan{ablation on layer}
% \haoyan{ablation on dictionary sizes}

\begin{rebuttalblock}
\subsection{Comparison with Post-hoc EMA Baselines}
\label{app:posthoc-ema}

The Persistent SAE learns three things together: which features to detect (the encoder), how long each feature persists ($\lambda_j$), and how to reconstruct the hidden state from the feature state (the decoder).
In this section, we compare our method with standard SAEs with fixed or post-hoc learned EMA, to check whether jointly learning feature-specific timescales is a necessary component.\looseness=-1

\myparagraph{Baselines.}
We take the trained Gemma-2-2B BatchTopK SAE (\Cref{app:sae-baselines}) and keep its encoder fixed, so each baseline has the same token-wise activations $\mathbf{a}^{(t)}$ as BatchTopK. We accumulate these activations into a state $\mathbf{h}^{(t)}$ with \Cref{eq:state-update} and reconstruct the hidden state from $\mathbf{h}^{(t)}$. We construct two types of baselines:
\begin{enumerate}[leftmargin=*,itemsep=1pt,topsep=2pt]
    \item \textbf{Post-hoc EMA with fixed $\lambda$}: all features share one $\lambda\in\{0.5,0.9,0.99\}$.
    \item \textbf{Post-hoc EMA with learned $\lambda_j$}: each feature learns its own $\lambda_j$ with the same parametrisation and initialisation (\Cref{app:persistence-param}). It differs from the Persistent SAE only in that the encoder is not trained together with the timescales.
\end{enumerate}

\myparagraph{Training.}
For both baseline types, we learn the $\lambda_j$ either with the original BatchTopK decoder fixed or with a refitted decoder. We fit the learned $\lambda_j$ and the refitted decoders on $890$k Pile tokens, in $256$-token sequences, for three epochs. For the features that never activate on these tokens, we set their $\lambda_j$ close to zero.

\myparagraph{Evaluation.}
For comparison, we measure FVE on the Pile dataset used in \Cref{tab:core-performance} and run sparse probing experiments using the setup in \Cref{app:probing-details}.
Since there's no timescale ranking in fixed-$\lambda$ baselines, we add their features in random order. For the learned-$\lambda_j$ baselines and the Persistent SAE, we add features in decreasing order of $\lambda_j$ (slowest first). We also report a random order for the Persistent SAE.

\begin{rebuttalblock}
\begin{table}[t]
\centering
\ifshowrebuttal\color{rebuttalcolor}\fi
\footnotesize
\setlength{\tabcolsep}{3pt}
\renewcommand{\arraystretch}{1.12}
\caption{
\textbf{Comparison to Post-hoc EMA baselines.}
On Gemma-2-2B, each post-hoc EMA baseline keeps the BatchTopK encoder fixed and averages its token-wise activations over time (\Cref{app:posthoc-ema}).
\emph{Ranking} gives the order in which we add features to the probe: \emph{Slowest} adds features in decreasing order of $\lambda_j$, and \emph{Random} averages three random orders ($\pm$ one standard deviation).
For fixed $\lambda$, FVE is shown as original\,/\,refitted decoder. Both decoders share one AUC, because the probe reads the state and the decoder does not change it.
Best/second-best results are bold/underlined.
}
\resizebox{\linewidth}{!}{%
\begin{tabular}{llcccccc}
\toprule
\multirow{2}{*}{Representation}
& \multirow{2}{*}{Ranking}
& \multirow{2}{*}{FVE ($\uparrow$)}
& \multicolumn{5}{c}{Sparse probing AUC ($\uparrow$) by feature fraction} \\
\cmidrule(lr){4-8}
& & & $0.5\%$ & $1\%$ & $5\%$ & $10\%$ & $100\%$ \\
\midrule
BatchTopK
& Random
& $\mathbf{0.73}$
& $0.52{\scriptscriptstyle\pm.00}$
& $0.52{\scriptscriptstyle\pm.00}$
& $0.58{\scriptscriptstyle\pm.01}$
& $0.62{\scriptscriptstyle\pm.01}$
& $0.79{\scriptscriptstyle\pm.00}$
\\
\quad + post-hoc EMA, fixed $\lambda=0.5$
& Random
& $0.58\,/\,\underline{0.66}$
& $0.59{\scriptscriptstyle\pm.01}$
& $0.60{\scriptscriptstyle\pm.01}$
& $0.65{\scriptscriptstyle\pm.00}$
& $0.69{\scriptscriptstyle\pm.01}$
& $0.85{\scriptscriptstyle\pm.00}$
\\
\quad + post-hoc EMA, fixed $\lambda=0.9$
& Random
& $0.25\,/\,0.33$
& $0.62{\scriptscriptstyle\pm.02}$
& $0.64{\scriptscriptstyle\pm.01}$
& $\underline{0.74{\scriptscriptstyle\pm.00}}$
& $\underline{0.78{\scriptscriptstyle\pm.01}}$
& $\mathbf{0.91{\scriptscriptstyle\pm.00}}$
\\
\quad + post-hoc EMA, fixed $\lambda=0.99$
& Random
& $0.09\,/\,0.12$
& $\underline{0.63{\scriptscriptstyle\pm.04}}$
& $\underline{0.65{\scriptscriptstyle\pm.03}}$
& $0.73{\scriptscriptstyle\pm.01}$
& $0.76{\scriptscriptstyle\pm.00}$
& $\underline{0.87{\scriptscriptstyle\pm.00}}$
\\
\quad + post-hoc EMA, learned $\lambda_j$, original decoder
& Slowest
& $\mathbf{0.73}$
& $0.54$
& $0.55$
& $0.55$
& $0.64$
& $0.81$
\\
\quad + post-hoc EMA, learned $\lambda_j$, refitted decoder
& Slowest
& $\mathbf{0.73}$
& $0.55$
& $0.56$
& $0.61$
& $0.65$
& $0.81$
\\
\midrule
\multirow{2}{*}{Persistent SAE (Ours)}
& Random
& \multirow{2}{*}{$\mathbf{0.73}$}
& $0.54{\scriptscriptstyle\pm.01}$
& $0.56{\scriptscriptstyle\pm.02}$
& $0.62{\scriptscriptstyle\pm.01}$
& $0.66{\scriptscriptstyle\pm.01}$
& $0.83{\scriptscriptstyle\pm.00}$
\\
& \cellcolor{gray!10}Slowest
&
& \cellcolor{gray!10}$\mathbf{0.79}$
& \cellcolor{gray!10}$\mathbf{0.80}$
& \cellcolor{gray!10}$\mathbf{0.81}$
& \cellcolor{gray!10}$\mathbf{0.81}$
& \cellcolor{gray!10}$0.83$
\\
\bottomrule
\end{tabular}%
}
\label{tab:posthoc-ema}
\end{table}
\end{rebuttalblock}

\myparagraph{Persistent SAE outperforms all post-hoc baselines and benefits from joint learning.}
In \Cref{tab:posthoc-ema}, we show that neither post-hoc baseline matches the Persistent SAE. 
For fixed $\lambda$ baselines, they lose much of the reconstruction quality. For learned $\lambda_j$ baselines, they preserve reconstruction quality, but their slowest features carry little subject information.
Since the encoder is fixed, a post-hoc EMA can only add memory to features learnt for single tokens. 
Only by jointly learning features and their timescales does the Persistent SAE achieve good reconstruction and concentrate subject information in a small set of slow features, clearly separating the roles of token-wise activations and the persistent state.
\end{rebuttalblock}

\begin{rebuttalblock}
\section{Cross-Feature Interactions of Persistent SAEs}
\label{app:cross-feature}

Persistent SAEs model temporal persistence separately for each feature, using a fixed, learned coefficient $\lambda_j$. This simple design gives every feature an interpretable timescale, but it does not explicitly model dependencies between features (\Cref{sec:discussion-conclusion}). However, in \Cref{sec:slow-features}, we show that the slowest features jointly track subject information, suggesting that the features may still interact meaningfully over time.
We therefore ask two follow-up questions: (1) whether, and how, features interact over time for the Persistent SAE, and (2) whether modelling interactions between features explicitly improves reconstruction quality.

\myparagraph{Setup.}
To answer both questions, we need to find a set of related features for each feature. We find them by measuring how often features activate near each other.
For each pair of features $i$ and $j$, we count how often $j$ activates within $W=16$ tokens of $i$, and divide this count by the count expected by chance. We average this ratio over both directions to obtain the co-activation ratio:
\begin{equation}
    r_{ij} =
    \frac{\text{observed co-activations}}
         {\text{expected co-activations}}.
    \label{eq:coactivation-ratio}
\end{equation}
A ratio of $1$ means the two features activate near each other as often as chance, and a larger ratio means they do so more often.
For each feature, we define the $K=8$ features with the highest $r_{ij}$ as its \textbf{partners}, and we call the feature and its partners a \textbf{feature group}. 
% We ignore pairs that co-activate only a few times, since their ratio is unreliable. 
As a control, we also pick $K$ random features for each feature that match its partners in how often they activate and in $\lambda_j$.
We compute all counts from the token-wise activations $\mathbf{a}^{(t)}$.
We find these partners on 2,048 Pile documents that we keep apart from every evaluation in this section, and we evaluate on 512 new Pile documents. 
We find feature partners separately for each language model (Pythia-160M and Gemma-2-2B) and each SAE: on the Persistent SAE in \Cref{app:group-interaction} and on the BatchTopK SAE in \Cref{app:cross-feature-correction}.

\subsection{Feature Interactions in Persistent SAEs Capture Topic-Level Information}
\label{app:group-interaction}

Although Persistent SAEs model persistence separately for each feature, features may still interact through the way they activate together. 
We therefore test two things: whether a feature and its partners share a topic, and whether they stay active together while the subject of the text continues.

\myparagraph{Co-activating features form topic groups.}
First, we compute the co-activation ratio $r_{ij}$ (\Cref{eq:coactivation-ratio}) on $512$ held-out Pile documents. We show three illustrative groups, each made of a feature and its $5$ strongest partners, chosen to cover different topics and to exclude features that activate on very common tokens.
In \Cref{fig:group-interaction}a, features in the same group activate near each other far more often than chance, whereas features from different groups rarely do, and each group corresponds to a specific topic (sports, clinical studies, and family).
This suggests that features on the same topic activate together, even though each feature persists on its own.

\myparagraph{Topic groups jointly track the subject over time.}
Next, we test whether these groups stay active together while the subject continues, using the partners and matched random features defined in \Cref{app:cross-feature}.
Following the spliced MMLU setup of \Cref{sec:slow-features}, we join $256$-token segments of MMLU questions from four subjects. Each segment is followed once by a new document on the same subject and once by a new document on a different subject. Both continuations start a new document, so only the subject differs.
For each feature that activates in the first segment, we measure the average activation of its partners, of the feature itself, and of its matched random features in windows of the new document ($0$--$16$, $16$--$64$, $64$--$128$, and $128$--$256$ tokens after the switch). We then divide the activation when the subject continues by that when the subject changes, so a ratio of $1$ means that the subject makes no difference.

\begin{figure}[t]
    \centering
    \includegraphics[width=\linewidth]{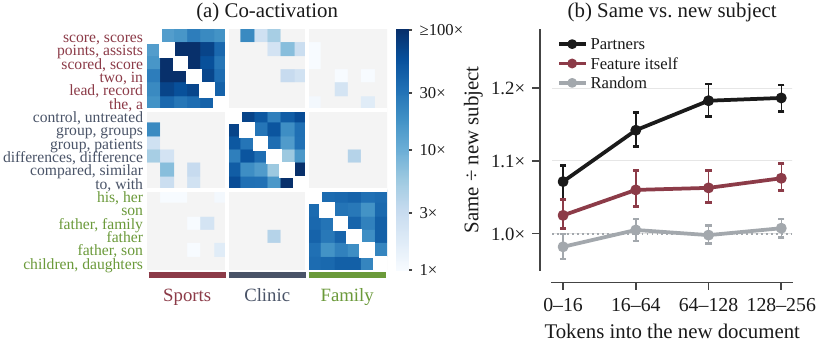}
    \caption{\protect\rebuttal{
        \textbf{Features of the Persistent SAE form topic groups that follow the subject.}
        \textbf{(a)}~Co-activation ratio $r_{ij}$ (\Cref{eq:coactivation-ratio}, log scale) for three illustrative groups of the Pythia-160M Persistent SAE. Each row is one feature, labelled by the words on which it most often activates.
        \textbf{(b)}~Average activation of a feature's partners, the feature itself, and matched random features when a new document continues the subject, divided by that when it changes subject, for the Gemma-2-2B Persistent SAE. Error bars are 95\% confidence intervals from resampling features.
    }}
    \label{fig:group-interaction}
\end{figure}

In \Cref{fig:group-interaction}b, a feature's partners are more active when the new document continues the same subject, and this gap lasts up to $256$ tokens into the new document. The feature itself shows a smaller gap, and matched random features show none. The partners therefore track the subject more strongly than the feature itself.

These results help explain why a small set of slow features tracks the subject in \Cref{sec:slow-features}. Each feature has its own $\lambda_j$, yet related features activate together and follow the subject as a group, so subject information is spread across groups of features rather than held by single features. 
Persistent SAEs therefore already show meaningful cross-feature interaction without explicit cross-feature terms.
This interaction emerges from features that each persist on their own. 
We next test whether modelling such interactions explicitly could further improve reconstruction quality.

\subsection{Does Cross-Feature Persistence Improve Reconstruction?}
\label{app:cross-feature-correction}

\myparagraph{Setup.}
To explore whether modelling these interactions explicitly helps an SAE reconstruct the hidden states better than single-feature persistence, we extend the context-correction experiment in \Cref{sec:temporal-persistence} (\Cref{eq:context-head}) for the standard BatchTopK SAE, on which we first observed temporal structure of SAE features.
Specifically, we keep the SAE fixed and add a correction to its reconstruction $\hat{\mathbf{x}}^{(t)}_{\mathrm{base}}$ that uses the token-wise activations from $\tau$ earlier tokens:
\begin{equation}
    \hat{\mathbf{x}}^{(t)}_{\mathrm{ctx}}
    =
    \hat{\mathbf{x}}^{(t)}_{\mathrm{base}}
    +
    \mathbf{W}_{\mathrm{dec}}
    \left(\mathbf{w}_\tau \odot \mathbf{a}^{(t-\tau)} + \mathbf{M}_\tau\, \mathbf{a}^{(t-\tau)}\right).
    \label{eq:cross-feature-head}
\end{equation}
The first term lets a past feature add back its own direction, as in \Cref{sec:temporal-persistence}. The matrix $\mathbf{M}_\tau$ lets a past feature also add back the directions of $8$ other features, with one learned weight per link. We compare four ways to choose these interacted features:
\begin{enumerate}[leftmargin=*,itemsep=1pt,topsep=2pt]
    \item \textbf{Random}: randomly selected features.
    \item \textbf{Co-activating partners}: its partners, the features that activate near it (\Cref{eq:coactivation-ratio}).
    \item \textbf{Random, partner-level overlap}: random features that do not co-activate with it, but whose decoder directions overlap with its own as much as those of its partners. Comparing them with the partners isolates the effect of co-activation.
    \item \textbf{Nearest directions}: the features whose decoder directions are closest to its own. Adding back these directions mainly refines the reconstruction along the feature's own direction.
    % , so this choice serves as a comparable approximation to single-feature modelling.
\end{enumerate}
All four choices have the same number of learned weights.
We fit the correction on $1{,}024$ Pile documents, stop training based on $128$ further documents, and measure $R^2_{\mathrm{improve}}$ (\Cref{eq:held-out-fraction-correction}) on $256$ test documents, with three random seeds.

\begin{figure}[t]
    \centering
    \includegraphics[width=0.85\linewidth]{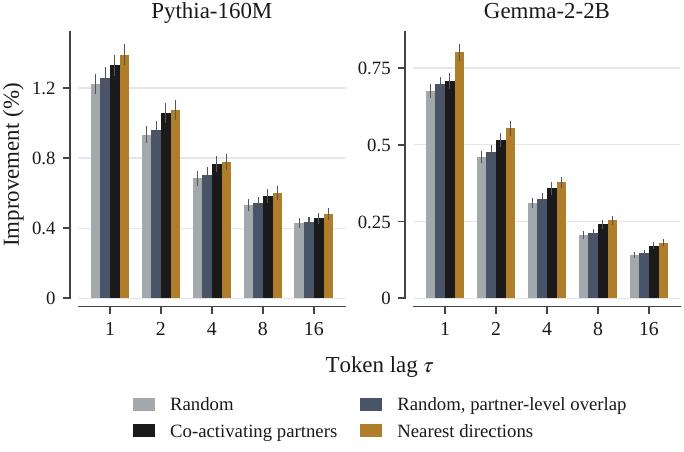}
    \caption{\protect\rebuttal{
        \textbf{Nearest direction help reconstruction better than nearest directions co-activating partners}
        Co-activating partners give a smaller gain in reconstruction ($R^2_{\mathrm{improve}}$) than nearest directions. Error bars are 95\% confidence intervals.
    }}
    \label{fig:cross-feature-correction}
\end{figure}

\myparagraph{Nearest directions help reconstruction more than co-activating partners.}
In \Cref{fig:cross-feature-correction}, interacting each past feature to its nearest directions improves reconstruction more than interacting it to its co-activating partners. Co-activating partners give only a small gain over random features with the same overlap. Contrary to our expectation, the past information that helps reconstruction therefore lies mainly along each feature's own direction. This supports the design of Persistent SAEs, which model persistence with one coefficient $\lambda_j$ per feature (\Cref{sec:method}). The small gain from co-activation still makes explicit cross-feature modelling a direction for future work, although reconstruction alone may be a weak signal to justify it.
% \myparagraph{Cross-feature interaction does not outperform single-feature modelling.}
% In \Cref{fig:cross-feature-correction}, linking each past feature to its co-activating partners does not improve reconstruction more than linking it to its nearest directions, which approximate single-feature modelling, and gives only a small gain over random features with the same overlap. The past information that helps reconstruction therefore lies mainly along each feature's own direction. This supports the design of Persistent SAEs (\Cref{sec:method}), which model persistence with one coefficient $\lambda_j$ per feature: an explicit cross-feature term would add parameters without a clear gain in reconstruction.
% Nevertheless, this small gain from co-activation suggests that explicit cross-feature modelling is a promising direction for future work, although reconstruction alone is a weak training signal, and such designs require further evidence to justify them.
\end{rebuttalblock}

\end{document}